\pdfoutput=1

\documentclass[11pt]{article}

\usepackage{EMNLP2022}

\usepackage{times}
\usepackage{latexsym}
\usepackage{microtype}
\usepackage{booktabs}
\usepackage{graphicx}
\usepackage{amsmath}
\usepackage{enumitem}
\usepackage{float}
\usepackage{multirow}
\usepackage[textsize=scriptsize]{todonotes}
\usepackage{dingbat}
\usepackage{url}

\usepackage{color, colortbl}

\usepackage[T1]{fontenc}

\usepackage[utf8]{inputenc}

\usepackage{microtype}

\usepackage{inconsolata}

\newcommand{\gpt}{\texttt{GPT3-D2}}
\newcommand{\brio}{\texttt{BRIO}}
\newcommand{\tzero}{\texttt{T0}}

\title{News Summarization and Evaluation in the Era of GPT-3}

\author{Tanya Goyal$^1$ \hspace{0.7cm} Junyi Jessy Li$^2$ \hspace{0.7cm} Greg Durrett$^1$ \\
    $^1$ Department of Computer Science \hspace{0.7cm}
    $^2$ Department of Linguistics \\
  The University of Texas at Austin \\
  {\tt tanyagoyal@utexas.edu}}

\begin{document}
\maketitle
\begin{abstract}
The recent success of prompting large language models like GPT-3 has led to a paradigm shift in NLP research. In this paper, we study its impact on text summarization, focusing on the classic benchmark domain of news summarization. First, we investigate how GPT-3 compares against fine-tuned models trained on large summarization datasets. We show that not only do humans overwhelmingly prefer GPT-3 summaries, prompted using only a task description, but these also do not suffer from common dataset-specific issues such as poor factuality. Next, we study what this means for evaluation, particularly the role of gold standard test sets. Our experiments show that both reference-based and reference-free automatic metrics cannot reliably evaluate GPT-3 summaries.
Finally, we evaluate models on a setting beyond generic summarization, specifically keyword-based summarization, and show how dominant fine-tuning approaches compare to prompting.

To support further research, we release: (a) a corpus of 10K generated summaries from fine-tuned and prompt-based models across 4 standard summarization benchmarks, (b) 1K human preference judgments comparing different systems for generic- and keyword-based summarization.\footnote{All data available at: \url{https://tagoyal.github.io/zeroshot-news-annotations.html}.}
\end{abstract}

\section{Introduction}
Fine-tuning pre-trained models on domain-specific datasets has been the leading paradigm in text summarization research in recent years \cite{lewis2020bart, zhang2020pegasus, raffel2020exploring}. These models generate high-quality summaries on standard benchmarks, but still require sizeable training datasets to adapt to new settings, e.g., summarizing data from a new source domain or producing a summary in a different style. 
The success of prompting large language models (GPT-3 \cite{brown2020language}, T0 \cite{sanh2022multitask}, PaLM \cite{chowdhery2022palm}, etc.) provides an alternative approach, namely learning from natural language task instructions and/or a few demonstrative examples in the context without updating model parameters. While recent work \cite{zhao2021calibrate, min2022rethinking, ye2022unreliability} has evaluated this paradigm across a number of tasks, it has only been studied for text summarization with unreliable automatic metrics \cite{he2022z, chowdhery2022palm, ouyang2022training} or in non-standard settings \cite{saunders2022self}.

\begin{figure}
    \centering
    \includegraphics[scale=0.236, trim=50mm 95mm 70mm 30mm, clip]{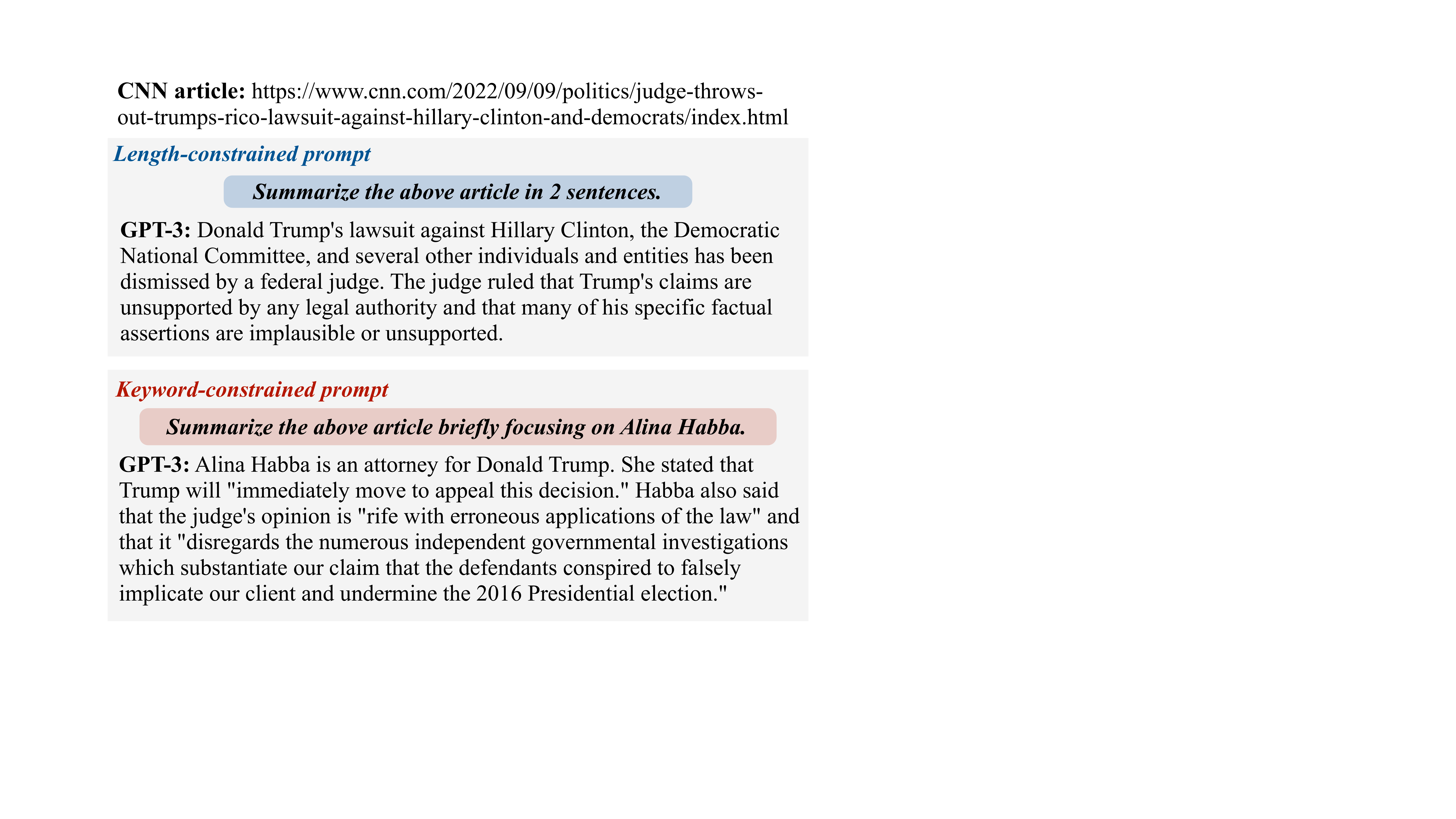}
    \caption{Examples of GPT-3 summaries. We can generate summaries following style constraints or queries included in the prompts, allowing us to emulate a range of existing fine-tuned systems.}
    \label{fig:length-keyword}
\end{figure}

In this paper, we conduct the first systematic study of the impact of prompt-based models on the text summarization research space, using an Instruct-tuned 175B GPT-3 model (text-davinci-002) \cite{brown2020language, ouyang2022training} as a case study. Figure~\ref{fig:length-keyword} shows that GPT-3 summaries are extremely high-quality and adaptable to different summarization settings. Starting from these observations, we aim to answer three main questions. First, how do prompt-based GPT-3 summaries compare to those obtained from state-of-the-art fine-tuned summarization models \cite{zhang2020pegasus, liu2022brio}?
We compare these approaches using A/B testing on a new corpus of recent news articles, and find that our study participants overwhelmingly prefer GPT-3 summaries across two different ``styles'' with different prompts (three-sentence and single-sentence). 
Moreover, these summaries do not suffer from limitations due to low-quality training data that plague fine-tuned generic summarization models \cite{maynez2020faithfulness, goyal2022training}.

Second, are existing automatic metrics well-suited to evaluating prompt-based summaries? Recent work has shown that classic reference-based such as \texttt{ROUGE} \cite{lin2004rouge} and \texttt{BERTScore} \cite{bertscore2020} are unreliable when small improvements are reported \cite{peyrard2019studying, fabbri2021summeval}; however large differences, on the order of say $5$ \textsc{Rouge} points or greater, are considered to be correlated with human preferences \cite{bhandari2020metrics, deutsch-etal-2022-examining}. However, we find that the same is no longer true when evaluating GPT-3 summaries. These summaries score much lower on automatic metrics ($7$ \texttt{ROUGE-L} points on average) than all prior state-of-the-art models while comfortably outperforming them on human evaluation. Furthermore, we show that recent reference-free metrics, e.g. QA-based metrics \cite{fabbri-etal-2022-qafacteval, durmus2020feqa} and trained factuality models \cite{kryscinski2020evaluating, goyal2020evaluating}, similarly fail to adapt to this shift from the fine-tuned to prompting, and need to be re-visited. 

Finally, how can prompting be used beyond generic summarization? We focus on keyword-based and aspect-based summarization. For keyword-based summarization, we find that GPT-3 consistently generates more coherent and keyword-relevant summaries compared to current fine-tuned alternatives: crowd annotators prefer GPT-3 summaries over a baseline model \cite{he2020ctrlsum} 70\% of the time. We observe mixed results for the aspect-based setting, where GPT-3 summaries show frequent failure cases with simple prompts.

Taken together, this evidence suggests that GPT-3 represents a fundamental paradigm shift in summarization, changing what data we need (or don't need) and what approaches we can now explore. Evaluating these systems will require a new framework distinct from the automatic metrics that have dominated the last decade of summarization research.

\section{Models and Setup}
\label{sec:background}

\subsection{Current Paradigms for Summarization}

Recent zero- and few-shot prompting based models \cite{brown2020language, sanh2022multitask}, have shown impressive generalization capabilities on unseen tasks specified using prompts alone and without performing any gradient updates \cite{mishra2022cross}. In this work, we want to compare their text summarization performance against the current state-of-the-art models.

\begin{figure}[h]
    \centering
    \includegraphics[scale=0.23, trim=25mm 190mm 40mm 40mm, clip]{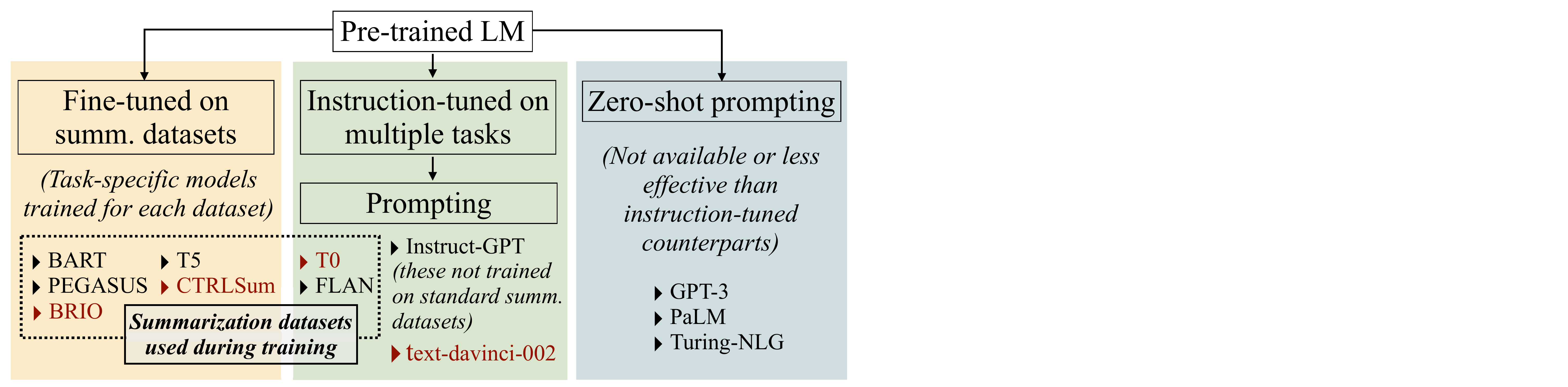}
    \caption{Broad categorization of available summarization systems; those compared in this work are highlighted in red.}
    \label{fig:landscape}
\end{figure}

Figure \ref{fig:landscape} shows the broad categories of all available summarization approaches, including current SOTA models and prompting-based models. The former set consists of \colorbox[HTML]{F7EBCC}{\textbf{fine-tuned}} language models, trained on a large number of article-summary pairs (e.g. BART \cite{lewis2020bart}, PEGASUS \cite{zhang2020pegasus}, BRIO \cite{liu2022brio}) to obtain dataset-specific systems. This category also includes models aimed at tasks beyond generic summarization, such as keyword- or query-based summarization, that still rely on standard datasets for training \cite{he2020ctrlsum}.

\begin{table}[t]
    \centering
    \small
    \begin{tabular}{c|cc|cc}
    \toprule
     \multirow{2}{*}{\textbf{Dataset}}   & \multicolumn{2}{c|}{\textbf{Avg. Words}} & \multicolumn{2}{c}{\textbf{\% novel n-grams}}  \\
        & Article & Summ & $n=1$ & $n=2$ \\ \midrule
        \textbf{\texttt{CNN}} & 760.5 & 45.7 & 16.7 & 54.3 \\
        \textbf{\texttt{DailyMail}} & 653.3 & 54.6 & 17.0 & 53.8 \\
        \textbf{\texttt{XSum (BBC)}} & 431.1 & 23.2 & 35.7 & 82.4 \\
        \textbf{\texttt{Newsroom}} & 658.6 & 26.7 & 18.9 & 47.5\\
    \bottomrule
    \end{tabular}
    \caption{Basic statistics of standard summarization datasets: \texttt{CNN/DM}~\cite{hermann2015teaching, nallapati2016abstractive}, \texttt{XSum} \cite{narayan2018don}, 
    Newsroom \cite{grusky2018newsroom}. These show large variance in their summary properties and fundamentally differ in their definition of the ``gold'' standard. }
    \label{tab:news_dataset_statistics}
\end{table}

On the other extreme are \colorbox[HTML]{D3DEE1}{\textbf{zero- or few-shot}} models, (e.g. GPT3 \cite{brown2020language}, PaLM \cite{chowdhery2022palm}), that are not explicitly trained for any particular task, as discussed above. Recent work \cite{ouyang2022training, wei2022finetuned, sanh2022multitask} has improved on these models by introducing \colorbox[HTML]{DCE5D2}{\textbf{instruction-tuned}} models. Here, pre-trained language models are fine-tuned on multiple tasks (which may include summarization) using instruction templates in order to align their training with inference time usage.

In this work, we compare the summarization performance of three  models that are representative of this space of options: 
\begin{enumerate}[leftmargin=*]
    \item \texttt{\textbf{OpenAI's text-davinci-002}}, a GPT-3 model \cite{brown2020language} from the Instruct series \cite{ouyang2022training}. While we do not know the exact training details for this release of the model, the previous in the series (text-davinci-001) was fine-tuned on a combination of prompts submitted to their API and labeler prompts spanning multiple tasks. These tasks include summarization but not (to our knowledge) standard summarization datasets like \texttt{CNN/DM}~\cite{hermann2015teaching, nallapati2016abstractive} or \texttt{XSum} \cite{narayan2018don}.
    
    We choose the text-davinci-002 version for our experiments in order to benchmark the best available prompt-based model.\footnote{We did not observe obvious quality differences in generated summaries between text-davinci-001 and text-davinci-002. Examples are included in Appendix~\ref{app:d1vsd2}.} We refer to this approach as \gpt{}. 
    
    \item \textbf{\brio} \cite{liu2022brio}, a fine-tuned summarization model that reports state-of-the art results on both \texttt{CNN/DM}~and \texttt{XSum}. We will use versions of this model fine-tuned on each of these two datasets.
    \item \textbf{\tzero}~\cite{sanh2022multitask}, a prompt-based model fine-tuned on multiple tasks including standard summarization datasets. This provides a useful point of comparison between task-specific fine-tuned (\brio) and bigger instruction-tuned models (\gpt).
\end{enumerate}

\subsection{Using \gpt~for summarization } Fine-tuned models largely follow the ``style'' of reference summaries in their training data, and hence, generated summaries show large variance between datasets (see Table \ref{tab:news_dataset_statistics} for basic summary statistics of standard summarization datasets). To ensure fair comparison between these and \gpt{}, we adapt the latter's prompt to align with dataset-specific styles.

\begin{figure}[t]
    \centering
    \includegraphics[scale=0.26, trim=60mm 110mm 25mm 50mm, clip]{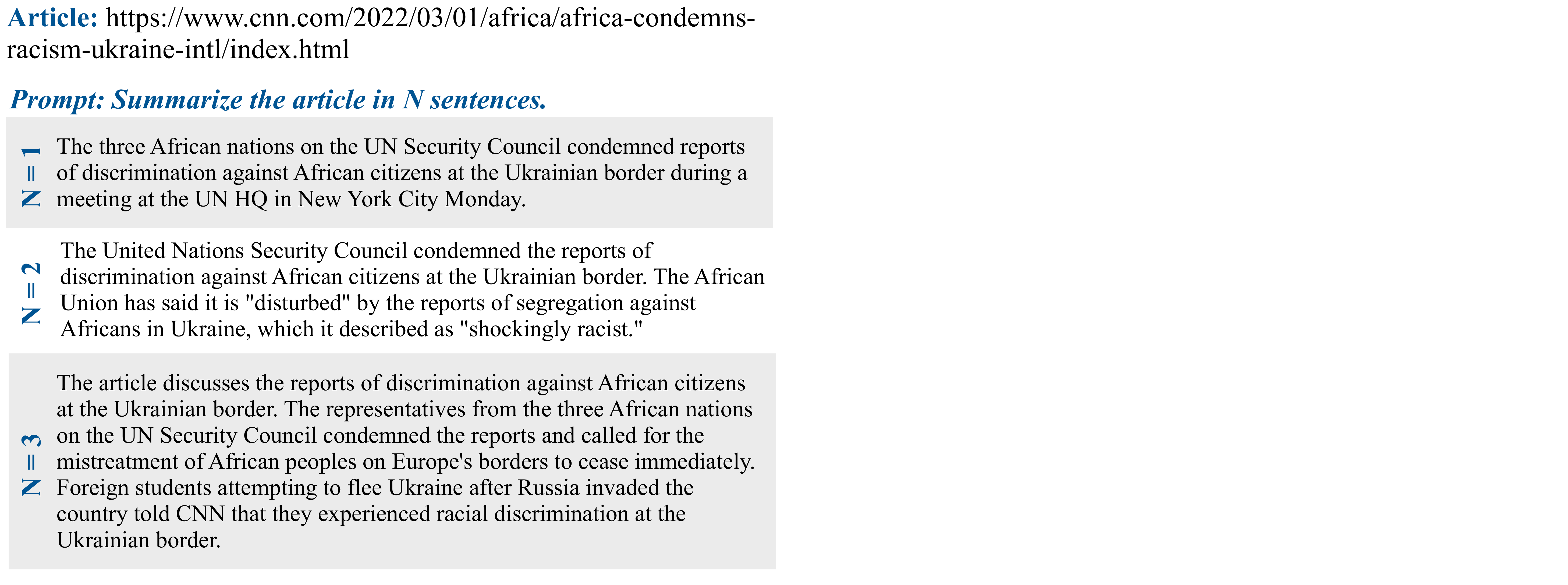}
    \caption{Illustration of length control using the task description / prompt for \gpt. We found that the generated summaries followed the given sentence length constraint 98\% of the time, allowing us to generate different length summaries emulating different datasets. }
    \label{fig:length_ex}
\end{figure}

Specifically, we follow prior work \cite{sanh2022multitask} and use sentence-count length prompts to adapt to each dataset. Although these datasets also differ along other attributes, e.g. \texttt{CNN/DM}~is lead-biased whereas \texttt{XSum} requires drawing inferences from a whole article, we do not attempt to control any other attributed of the summary. Figure~\ref{fig:length_ex} shows an example of different length \gpt~summaries for the same news article, using the following prompt format:
    
\paragraph{}\textit{Article: \{\{article\}\}}

\textit{Summarize the above article in {{N}} sentences.}
    
\paragraph{}We found that \gpt~summaries faithfully follow the given length constraint in 98\% of the test instances used in our human study data in Section~\ref{sec:human-study}.

Given this setup, we first compare the summary quality of the three summarization models through a human annotation study (Section \ref{sec:human-study}). Then, we evaluate the current suite of summarization metrics for prompt-based summarization (Section \ref{sec:automatic-eval}). Finally, in Section \ref{sec:keyword}, we briefly discuss \gpt~performance on summarization tasks beyond generic summarization and new challenges.

\begin{figure*}[t]
    \centering
    \includegraphics[width=\textwidth, trim=10mm 50mm 10mm 10mm, clip]{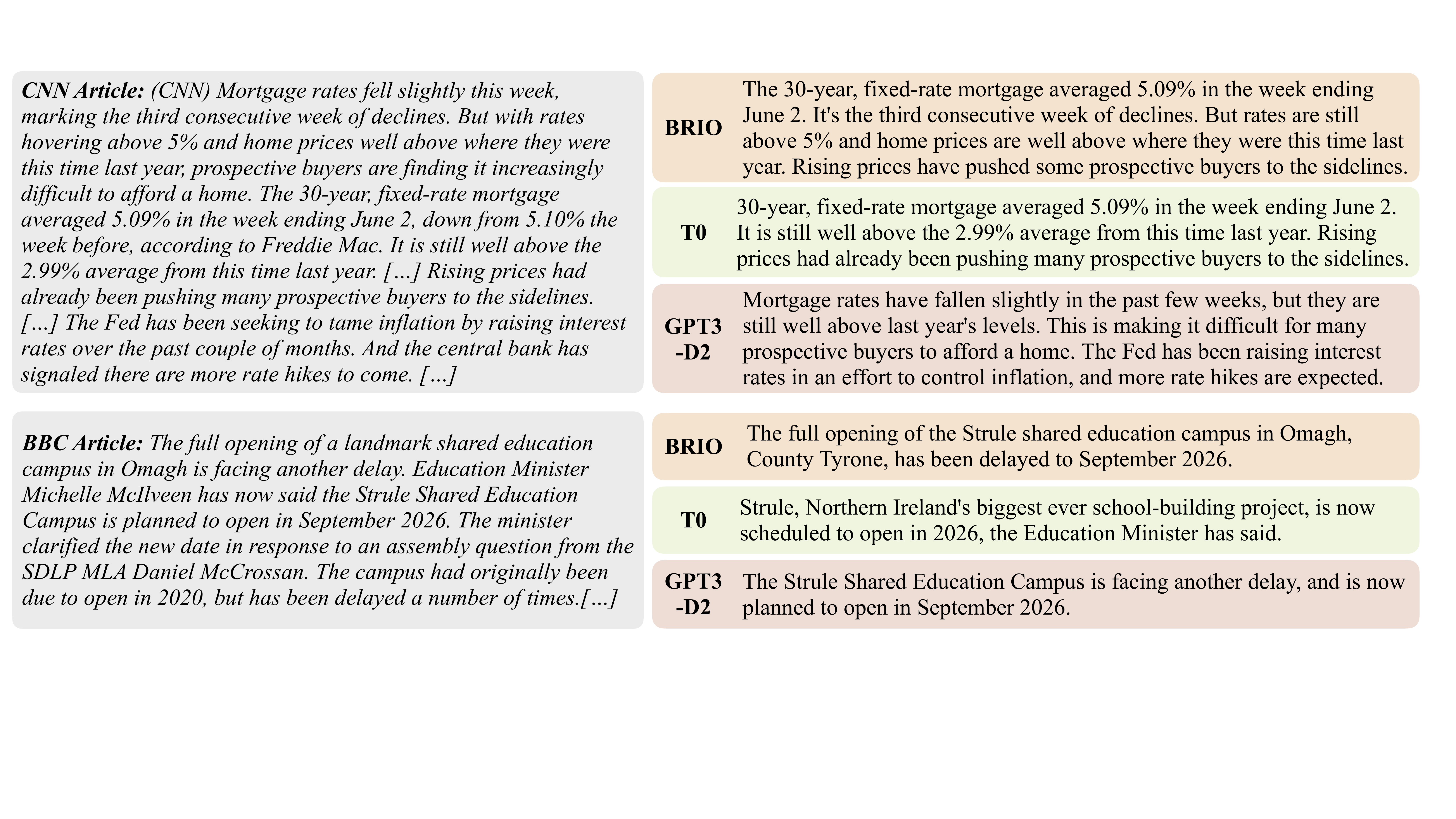}
    \caption{Examples of CNN-style and BBC/XSum-style summaries for the three systems. For CNN, we observe that models fine-tuned on the \texttt{CNN/DM}~training set reflect its dataset biases; summaries are highly extractive, specific and lead-biased. On the other hand, \gpt~summaries contain fewer specific details but cover more content.}
    \label{fig:qualitative_ex}
\end{figure*}

\section{Human evaluation of \gpt{} summaries}
\label{sec:human-study}

Generated summaries of fine-tuned models \cite{lewis2020bart, zhang2020pegasus, liu2022brio} emulate gold-standard summaries in their training datasets. In contrast, prompt-based \gpt~models generate summaries based on how the given task description surfaces behavior learned during pre-training or instruction-tuning. In this section, we ask: how do these paradigms compare? Does learning from gold summaries lead to a better summarization model? To answer this, we conduct a human study to compare  outputs of our 3 representative models and collect human preferences of quality.

\subsection{Experimental Setup}
\paragraph{Datasets for fine-tuning} We choose two standard fine-tuning datasets whose summaries differ along multiple dimensions such as length and abstractiveness: 

\begin{enumerate}[leftmargin=*]
    \item \textbf{CNN/DM} \cite{hermann2015teaching, nallapati2016abstractive} contains reference summaries that are approximately 3-4 sentences long. Summaries in this dataset are highly extractive and lead-biased.
    \item \textbf{XSum} \cite{narayan2018don} contains 1 sentence summaries of BBC news articles. In this dataset, references summaries, and consequently generated summaries from fine-tuned models are highly abstractive.
\end{enumerate}

\paragraph{Datasets for evaluation} Because \gpt{}'s pre-training and instruction-tuning datasets are unknown, it may have been trained on existing articles and summaries in the test splits of these standard benchmarks. 
We therefore run our human study on 100 recent articles from CNN\footnote{Although the \brio's \texttt{CNN/DM}~model also includes DailyMail data in its training, we do not use this news source in our study as it is now widely considered to be unreliable. E.g. according to Media Bias / Fact Check site, \texttt{DM}'s factual reporting is rated `low' \url{https://mediabiasfactcheck.com/daily-mail/}. } and BBC, collected between March 1, 2022 and June 31, 2022. We call these CNN-2022 and BBC-2022 respectively.

\paragraph{Model details} We use the publicly released \texttt{BRIO-XSum} and \texttt{BRIO-CNN/DM} models to generate summaries.\footnote{Models at: \url{https://github.com/yixinL7/BRIO}} For \tzero, we use a prompt we selected from its prompt repository for \texttt{CNN/DM}~and \texttt{XSum} datasets.\footnote{Repository with T0 prompts: \url{https://github.com/bigscience-workshop/promptsource}} Finally, to generate \gpt~summaries, we set $N=3$ for CNN and $N=1$ for BBC in our standard sentence-count prompt template from Section~\ref{sec:background}.

For a maximally fair comparison in this ``realistic'' setting, we take some additional steps to improve the output of \texttt{BRIO-XSum}. In order to automate dataset creation, \texttt{XSum} removes the first sentence from news articles to use as the gold summary for training, then treats the rest of the sentences as the article to summarize. This setup differs from the real world usage of summarization systems where the complete article is summarized. Due to this mismatch, \texttt{BRIO-XSum} often generates very low quality outputs, e.g. \textit{All images: Strule Shared Education Campus} in Figure \ref{fig:qualitative_ex}, for around 30\% of the articles. We manually identify these examples and first attempt to fix them by selecting a summary without such obvious failures from further down the beam (we use beam size $=10$). However, if we cannot find a ``better'' summary, we remove the first sentence of the article and re-sample a new summary to align with its noisy training. This latter strategy often results in factually incorrect summary generations, as is well documented in prior research \cite{maynez2020faithfulness, goyal2021annotating}.  

\paragraph{Design of the human study} We design an A/B test to collect preference annotations. For each given article, annotators are shown summaries from all three summarization systems (\brio, \tzero~and \gpt{}). They are then asked to select their most and least preferred summary or summaries. In addition to these multiple choice questions, we also ask for a free-text justification of both choices.

We make two design decisions for our human study: first, we do not provide annotators with specific definitions of summary quality to avoid introducing our own biases. It is also quite challenging to produce a unified definition of quality for the very different ``styles'' of summaries evaluated in this study.  Instead, we ask them to rely on their own preferences based on summaries they would like to see if they were browsing the web, which we believe to be a representative scenario for non-expert consumers of news summaries. Detailed task instructions are included in Appendix~\ref{app:instructions}.

Second, we allow multiple selections for both the best and worst summary questions to cater to scenarios in which different summarization systems output similar quality summaries without meaningful differences. 

We hire crowd annotators through Prolific. For both CNN and BBC, we recruit 60 unique participants to annotate the 100 summaries in each dataset. Each annotator was asked to annotate 5 articles and each article was annotated by 3 annotators. Additionally, we use the Prolific's demographic filters to restrict participation to USA (or UK) residents for CNN (or BBC). We anticipate that residents from these respective countries are better positioned to understand country-specific news events and evaluate their summaries. Participants were paid approximately \$11/hr for their work.

\begin{table}[t]
    \centering
    \small
    \renewcommand{\tabcolsep}{1mm}
    \begin{tabular}{r|cc|cc|c}
    \toprule
      \multirow{2}{*}{\textbf{Model}}  & \multicolumn{2}{c|}{\textbf{Length Statistics}}  & \multicolumn{2}{c|}{\textbf{\% novel n-gms}} & \textbf{\#NEs per} \\
       & \#sent & \#words/sent & $n=1$ & $n=2$ & \textbf{100 words} \\ \midrule
       \multicolumn{6}{c}{\textbf{CNN}} \\ \midrule
       \brio{} & 3.7 & 15.8 & 12.1 & 36.2 & 12.9 \\
       \tzero{} & 2.7 & 14.9 & 16.4 & 45.2 & 12.8 \\
       \gpt{} & 2.9 & 23.4 & 16.3 & 40.7 & 10.5 \\ \midrule
       \multicolumn{6}{c}{\textbf{BBC}} \\ \midrule
       \brio{} & 1.0 & 20.2 & 24.6 & 61.2 & 9.1\\
       \tzero{} & 1.0 & 20.0 & 26.3 & 66.7 & 9.8 \\
       \gpt{} & 1.0 & 27.7 & 16.4 & 42.3 & 8.5 \\ 
    \bottomrule
    \end{tabular}
    \caption{Statistics for generated summaries evaluated in the human study across all  datasets and summarization systems. We observe that \gpt{} generated summaries nearly always follow the sentence length constraints in their prompts.}
    \label{tab:quantitative}
\end{table}

\subsection{Results}

\paragraph{Differences between summarization systems} 
Figure \ref{fig:qualitative_ex} shows examples of generated summaries from all three summarization systems for both CNN and BBC articles. For CNN, we observe that fine-tuned \brio~summaries tend to be highly extractive and generally include a high number of named entities (dates, percentages, names), reflecting the data it was trained on. In contrast, \gpt~summaries are more abstractive and less specific, but provide a more exhaustive overview of the article content. Table~\ref{tab:quantitative} provides quantitative evidence of this; we use percentage of novel n-grams to measure abstractiveness, and number of named entities per 100 words to measure specificity. 

For BBC, we observe inverse trends where \brio~and \tzero~are more abstractive compared to \gpt. Again, this can be attributed to the \texttt{XSum} training data used to train both these prior models. For \gpt~summaries, on the other hand, the level of abstractiveness does not differ between datasets. Finally, Table~\ref{tab:quantitative} shows that \gpt~summaries tend to have longer sentences, and therefore similar number of summary sentences often results in a longer summary for both datasets. We study the effect of this length difference on human preference judgments in Appendix~\ref{app:length}.

\begin{table}[t]
\renewcommand{\tabcolsep}{1mm}
    \centering
    \small
    \begin{tabular}{c|cc|cc|cc}
    \toprule
        \multirow{2}{*}{\textbf{Dataset}} & \multicolumn{2}{c|}{\textsc{\textbf{Brio}}} & \multicolumn{2}{c|}{\textsc{\textbf{T0}}} & \multicolumn{2}{c}{\textsc{\textbf{GPT3}}} \\
        & Best $\uparrow$ & Worst $\downarrow$ & Best $\uparrow$ & Worst $\downarrow$ & Best $\uparrow$ & Worst $\downarrow$ \\ \midrule
        \textsc{\textbf{CNN}} & 36 & 24 & \cellcolor{red!20}8 & \cellcolor{red!20}67 & \cellcolor{green!20} 58 &  \cellcolor{green!20}9 \\
        \textsc{\textbf{BBC}} & \cellcolor{red!20}20 & \cellcolor{red!20}56 & 30 & 29 & \cellcolor{green!20} 57 &  \cellcolor{green!20}15 \\
    \bottomrule
    \end{tabular}
    \caption{Percentage of times a summarization system is selected as the best or worst according to majority vote (may be tied). Human annotators have a clear preference for \gpt{} for both CNN and BBC style summaries.}
    \label{tab:human-study-vanilla}
\end{table}

\paragraph{Which systems do humans prefer?} Results of our human study are summarized in Table \ref{tab:human-study-vanilla}. We report the percentage of times a particular system is the most/least preferred model according to majority vote combining all three annotator's choices.\footnote{As we allow multiple system selections, note that more that one system could be the majority. However, this is rare after majority vote: only 2\% of the articles in CNN and 7\% in BBC have multiple best summaries.}
Across both datasets and styles, we observe a clear preference for \gpt~summaries compared to the other two models. In fact, in both scenarios, the \gpt~outperforms the next best model by at least 20 percentage points. This improvement is statistically significant according to a paired bootstrap test (CNN $p-$value $= 2 \times 10^{-3}$, BBC $p-$value $= 6 \times 10^{-4}$). 

Note that the next best model differs between the two datasets. For BBC, annotators prefer \tzero~summaries over \brio. Annotator rationales often mentioned misleading or incorrect information as the primarily reason for selecting \brio~as the worst summary, confirming the issues that have been observed with \texttt{XSum}-trained models \cite{maynez2020faithfulness, pagnoni2021understanding, goyal2021annotating}. Although \tzero~also includes \texttt{XSum} training data, we hypothesize that its multi-task framework helps offset the noisy signal from \texttt{XSum}. 

In contrast, annotators rate \tzero~as the worst summarization system for CNN. The most common rationales for these were shorter length and inclusion of irrelevant details, e.g. long quotes, while missing key points. Some annotators also commented that these  \tzero~summaries were less coherent compared to the other models. Interestingly, we did not observe similar complaints for the single-sentence \tzero{} summaries for BBC.

\begin{figure}[t]
    \centering
    \includegraphics[scale=0.18, trim=48mm 80mm 40mm 20mm, clip]{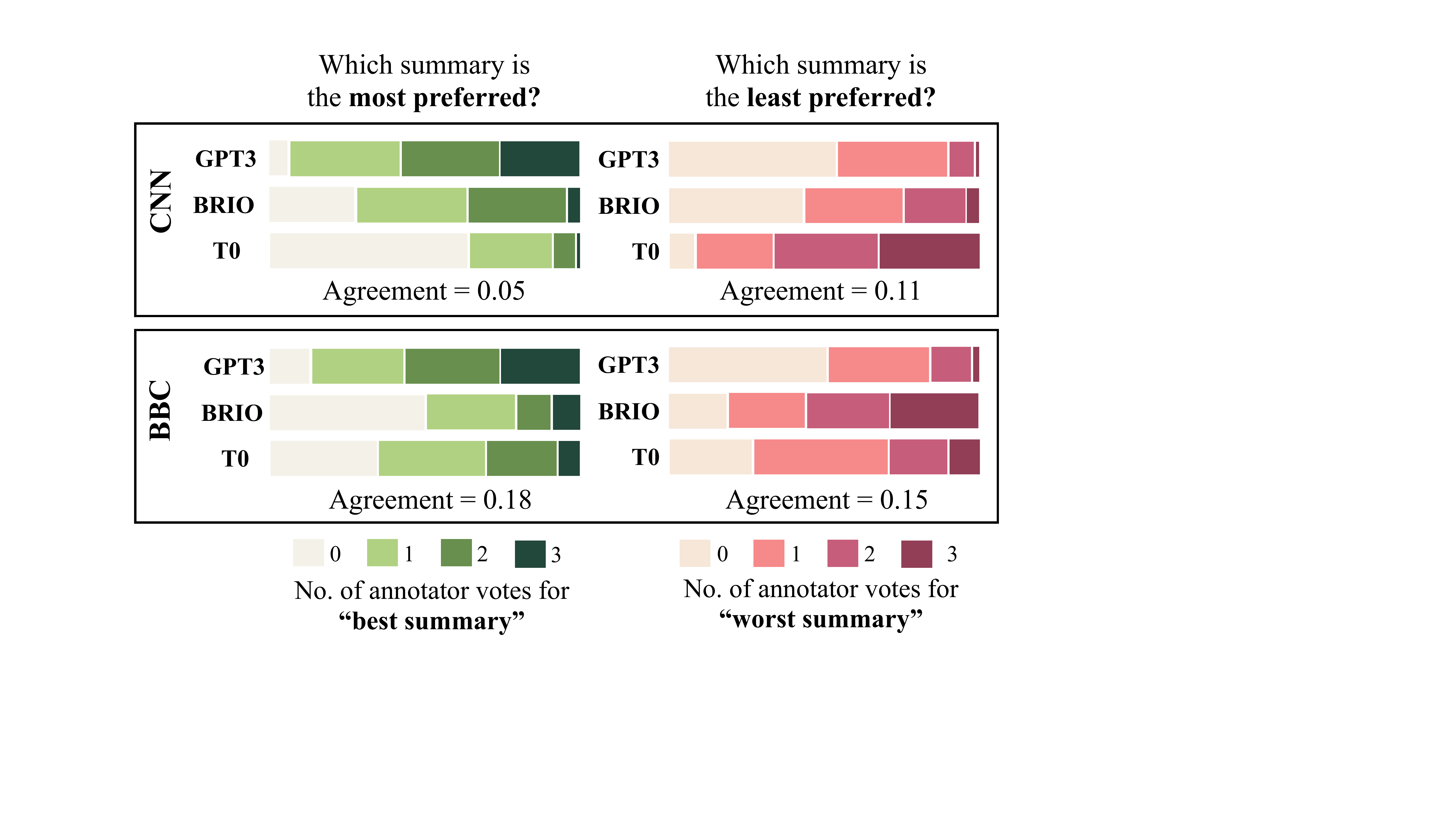}
    \caption{Annotator vote distribution for best and worst summaries across all datasets and models. Although \gpt~is the clear winner according to majority vote, this choice is unanimous for less than 30\% of the articles. This demonstrates the inherent variance in different annotators' definitions of ``best summary'', especially when comparing high-quality summaries from strong models. }
    \label{fig:vote_share}
\end{figure}

\paragraph{Do annotators agree with each other?} To study this, we plot the distribution of annotator votes for each summarization system and dataset in Figure~\ref{fig:vote_share}. Additionally, we report the inter-annotator agreement, measured using Krippendorff's alpha with MASI distance \cite{passonneau2006measuring}, to account for multiple selections of best or worst summary allowed in our study design.

The vote distribution shows that although more annotators prefer \gpt~summaries, this choice is only unanimous, i.e. supported by all three annotators, for less that 30\% of the annotated articles. Conversely, although \brio~(or \tzero) summaries are less preferred than \gpt~for the CNN (or BBC) dataset on aggregate, they were voted as the best summary by at least one annotator for more than 60\% of the articles. This demonstrate two things: first, when comparing summaries from two strong models, the choice is inherently ambiguous (similar observations in \citet{clark2021all}). Second, these results and the diversity in the written rationales, show that there does not exist a universal definition of a ``good'' summary and that different summary properties appeal to different annotators. Regardless, the aggregate preference for \gpt{} is high enough across the board to give us confidence in its strength.

\begin{table*}[t]
    \centering
    \small
    \begin{tabular}{c|c|ccc|cc|cc}
    \toprule
       \multirow{2}{*}{\textbf{Dataset}} & \multirow{2}{*}{\textbf{Model}} & \multicolumn{3}{c|}{\textbf{Overlap-Based}} & \multicolumn{2}{c|}{\textbf{Similarity-Based}} & \multicolumn{2}{c}{\textbf{QAEval}} \\
        & & \textbf{\textsc{ROUGE(1/2/L)}} & \textsc{\textbf{Meteor}} & \textsc{\textbf{Bleu}} & \textbf{BERTScore} & \textbf{MoverScore} & \textbf{EM} & \textbf{F1} \\ \midrule
        \multirow{4}{*}{\texttt{\textbf{CNN}}} 
        & \texttt{PEGASUS} & 34.85/14.62/28.23 & .24 & \cellcolor{green!20}{7.1} & \cellcolor{red!20}.858 & .229 & .105 & .160 \\
        & \brio{}  & \cellcolor{green!20}38.49/17.08/31.44 & \cellcolor{green!20}{.31} & 6.6 & \cellcolor{green!20}{.864} & \cellcolor{green!20}.261 & \cellcolor{green!20}.137 & \cellcolor{green!20}.211 \\
        & \tzero{}  & 35.06/13.84/28.46 & .25 & 5.9 & .859 & .238  & .099 & .163 \\ 
        & \gpt{}   & \cellcolor{red!20} 31.86/11.31/24.71 & \cellcolor{red!20}.25 & \cellcolor{red!20} 3.8 & \cellcolor{red!20}.858 & \cellcolor{red!20}.216 & \cellcolor{red!20}.098 & \cellcolor{red!20}.159 \\\midrule
        \multirow{4}{*}{\texttt{\textbf{DailyMail}}} 
        & \texttt{PEGASUS} & 45.77/23.00/36.65 & .33 & \cellcolor{green!20}12.2 & .865 & .308 & .159 & .229 \\
        & \brio{}  & \cellcolor{green!20}{49.27/24.76/39.21} & \cellcolor{green!20}{.37} & 11.7 & \cellcolor{green!20}.871 & \cellcolor{green!20}.331 & \cellcolor{green!20}.175 & \cellcolor{green!20}.259 \\
        & \tzero{} & 42.97/19.04/33.95 & .28 & 8.9 & .863 & .290 & .121 & .184 \\ 
        & \gpt{}  & \cellcolor{red!20} 38.68/14.24/28.08 & \cellcolor{red!20}.26 & \cellcolor{red!20}6.6 & \cellcolor{red!20}.859 & \cellcolor{red!20}.248 & \cellcolor{red!20}.101 & \cellcolor{red!20}.159 \\\midrule
        \multirow{4}{*}{\texttt{\textbf{XSum}}} 
        & \texttt{PEGASUS} & 47.97/24.82/39.63 & .36 & 9.8 & \cellcolor{green!20}{.901} & .362 & \cellcolor{green!20}.145 & .221 \\
        & \brio{}  & \cellcolor{green!20}{49.66/25.97/41.04} & \cellcolor{green!20}{.39} & \cellcolor{green!20}{10.6} & \cellcolor{green!20}{.901} & \cellcolor{green!20}.372 & .139 & \cellcolor{green!20}.224 \\
        & \tzero{}  &  44.20/20.72/35.84 & .34 & 8.0 & .896 & .340  & .125 & .208\\ 
        & \gpt{}   & \cellcolor{red!20}28.78/7.64/20.60 & \cellcolor{red!20}.19 & \cellcolor{red!20}2.2 & \cellcolor{red!20}.869 & \cellcolor{red!20}.197 & \cellcolor{red!20}.066 & \cellcolor{red!20}.119 \\\midrule
        \multirow{4}{*}{\texttt{\textbf{Newsroom}}} 
        & \texttt{PEGASUS} &  \cellcolor{green!20}39.21/27.73/35.68 &  \cellcolor{green!20}.39 &  \cellcolor{green!20}.14 & \cellcolor{green!20}.873 & \cellcolor{green!20}.272 & \cellcolor{green!20}0.182 & \cellcolor{green!20}0.253 \\
        & \brio{} & - & - & - & - & - & - & - \\
        & \tzero{}  & \cellcolor{red!20}25.64/9.49/21.41 & \cellcolor{red!20}.20 & \cellcolor{red!20}.04 & \cellcolor{red!20}.849 & \cellcolor{red!20}.145 & \cellcolor{red!20}.080 & \cellcolor{red!20}0.125 \\ 
        & \gpt{}   & 27.44/10.67/22.18 & .22 & .05 & .859 & .159 & .089 & 0.142 \\
    \bottomrule
    \end{tabular}
    \caption{Performance of different summarization systems measured using reference-based automatic metrics. Across all datasets, we observe that automatic metrics report substantially worse results for \gpt~summaries compared to fine-tuned models. This directly contradicts the human preference results from Section \ref{sec:human-study}, demonstrating that these reference-based metrics cannot reliably compare the quality of prompt-based summaries against fine-tuned summaries.}
    \label{tab:reference-based-automatic}
\end{table*}

\paragraph{How do these results impact the field?} Progress in text summarization research in the last five years has been enabled by the construction of large-scale text summarization datasets that involved scraping news articles and pairing them with any available summary-like data \cite{hermann2015teaching, narayan2018don, grusky2018newsroom}. The \texttt{CNN/DM}~dataset considers bullet points accompanying news articles as its summary. These ``gold'' standard summaries provided useful training signal to train impressive supervised models \cite{lewis2020bart, zhang2020pegasus, liu2022brio} and hence, their quality or alignment with human preferences was largely ignored. 

We found that, despite its popularity, \texttt{XSum} is largely unsuitable for fine-tuning models like \brio{} for realistic summarization settings. 
Even though a \texttt{CNN/DM}-trained \brio{} model performed better, the results of our human study question the continued utility of hill-climbing on this dataset, as it seems users may simply prefer a different style of summary altogether. In fact, this preference for \gpt{} is much larger than incremental improvements reported in other human evaluation settings, e.g. improvements on XSum on the \texttt{GENIE} leaderboard \cite{Khashabi2021GENIEAL}. Furthermore, as we we will see in Section~\ref{sec:keyword}, the greater flexibility of \gpt{} compared to these systems makes it more suitable for news summarization tasks beyond generic summarization.

If a system designer collects a large-scale dataset of high-quality summaries that they wish to emulate, we believe a fine-tuned system may outperform \gpt{}. However, better-trained models on datasets collected via ``incidental'' supervision are less likely to help.

\section{Can current automatic metrics evaluate \gpt~summaries?}
\label{sec:automatic-eval}

Automatic metrics proposed for summarization evaluation can be broadly divided into two categories: (1) \textbf{reference-based}, that compare generated summaries against available gold summaries, and (2) \textbf{reference-free} that only rely on the input document. Here, we compare their performance at evaluating prompt-based \gpt~summaries.

\begin{table*}[t]
    \centering
    \small
    \begin{tabular}{c|c|cc|cc|ccc}
    \toprule
        \multirow{2}{*}{\textbf{Dataset}} & \multirow{2}{*}{\textbf{Model}} & \multicolumn{2}{c|}{\textbf{Overall Quality}} & \multicolumn{2}{c|}{\textbf{Factuality (QA-based)}} & \multicolumn{3}{c}{\textbf{Factuality (NLI-based)}} \\
         & & \textbf{SUPERT} & \textbf{BLANC} & \textbf{QuestEval} & \textbf{QAFactEval} & \textbf{FactCC} & \textbf{DAE} & \textbf{SummaC} \\ \midrule
        \multirow{4}{*}{\texttt{\textbf{CNN}}}
        & \texttt{PEGASUS} & .5466 & .0605 & .7373 & \cellcolor{green!20}4.4071 & \cellcolor{green!20}.3743 & \cellcolor{green!20}.8223 & \cellcolor{green!20}.1138 \\
        & \brio{}  & \cellcolor{green!20}.5586 & \cellcolor{green!20}.0802 & .7334 & 3.8332 & \cellcolor{red!20}.1817 &  .7577 & -.0532 \\
        & \tzero{}  & \cellcolor{red!20}.5330 & \cellcolor{red!20}.0558 & \cellcolor{green!20}.7799 & 3.7517 & .2012 & .7556 & -.0605 \\ 
        & \gpt{}   & .5560 & .0749 & \cellcolor{red!20}.7249 & \cellcolor{red!20}3.6399 & .2428 & \cellcolor{red!20}.6671 & \cellcolor{red!20}-.0729 \\\midrule
        \multirow{4}{*}{\texttt{\textbf{DailyMail}}}
        & \texttt{PEGASUS} & \cellcolor{green!20}.6433 & .1137 & .7536 & \cellcolor{green!20}4.4677 & \cellcolor{green!20}.5152 & \cellcolor{green!20}.8497 & \cellcolor{green!20}.2402 \\
        & \brio{}  & .6360 & \cellcolor{green!20}.1217 & \cellcolor{red!20}.7415 & 4.1362 & .3699 & .8118 & \cellcolor{red!20}.0153\\
        & \tzero{}  & \cellcolor{red!20}.5995 & \cellcolor{red!20}.0889 & \cellcolor{green!20}.7803 & 3.9827 & \cellcolor{red!20}.2431 & .8043 & .0478\\ 
        & \gpt{}   & .6118 & .0983 & .7461 & \cellcolor{red!20}3.8279 & .2697 & \cellcolor{red!20}.6990 & .0365 \\\midrule
        \multirow{4}{*}{\texttt{\textbf{XSum}}}
        & \texttt{PEGASUS} & \cellcolor{red!20}.4439 & .0249 & .8233  & 2.0089 & .2465 & \cellcolor{red!20}.3598 & -.2993\\
        & \brio{} & .4459 & \cellcolor{red!20}.0230 & \cellcolor{green!20}.8305  & \cellcolor{red!20}1.8626 & \cellcolor{red!20}.2031 & .3040 & \cellcolor{red!20}-.3292\\
        & \tzero{}  & .4538 & .0238 & \cellcolor{red!20}.7957  & 2.0330 & .2219 & .3392 & -.3037\\ 
        & \gpt{}   & \cellcolor{green!20}.5060 & \cellcolor{green!20}.0594 & .8064  & \cellcolor{green!20}2.9492 & \cellcolor{green!20}.3977 & \cellcolor{green!20}.6372 & \cellcolor{green!20}-.2626\\\midrule
        \multirow{4}{*}{\texttt{\textbf{Newsroom}}}
        & \texttt{PEGASUS} & \cellcolor{green!20}.6286 & \cellcolor{green!20}.1131 & \cellcolor{red!20}.7118  & \cellcolor{green!20}4.2120 & \cellcolor{green!20}.7218 & \cellcolor{green!20}.7956 & \cellcolor{green!20}.2418 \\
        & \brio{}  & -&-&-&-&-&-&- \\
        & \tzero{}  & .5433 & .0640 & \cellcolor{green!20}.7511  & 3.5799 & \cellcolor{red!20}.2828 & .7376 & .0261 \\ 
        & \gpt{}   & \cellcolor{red!20}.5408 & \cellcolor{red!20}.0599 & .7160 & \cellcolor{red!20}3.2336 & .3988 & \cellcolor{red!20}.6564 & \cellcolor{red!20}-.0729\\
    \bottomrule
    \end{tabular}
    \caption{Performance of different summarization systems, as scored by automatic reference-free evaluation metrics from the summarization literature. Similar to reference-based metrics, these also generally fail to produce the same system rankings as human preferences reliably across datasets.}
    \label{tab:reference-free}
\end{table*}

\paragraph{Experimental Setup} We evaluate automatic metrics using summaries from 4 different summarization datasets, listed in Table~\ref{tab:news_dataset_statistics}. For each dataset, we construct our evaluation sets by randomly sampling 500\footnote{This size is chosen to give sufficient statistical power \cite{card2020little} while keeping costs for \gpt{} evaluation low to enable others to compare on this subset. We outline costs in Appendix \ref{sec:cost}.} articles from the standard test split.\footnote{Note that these standard datasets were released before 2020. Therefore, it is possible that some article-summary pairs in our test set overlap with \gpt's training data. However, we do not observe a qualitative difference in \gpt{}'s performance on these older articles.} We compare the same 3 summarization systems from Section \ref{sec:human-study} in our analysis. Additionally, we also report results using the fine-tuned \texttt{PEGASUS} model \cite{zhang2020pegasus}, as \brio~fine-tuned models are not available for all datasets.

We publicly release this corpus of summarization outputs to standardize the test sets and support future research into \gpt{} based summarization. Link: \url{https://tagoyal.github.io/zeroshot-news-annotations.html}.

\subsection{Reference-based metrics}
\label{sec:reference-based}

Here, we study if the gold summaries of the standard datasets are useful for evaluation, especially when evaluating prompt-based summaries that are not trained to emulate the gold. We benchmark the performance of 3 different summarization metrics: (1) \textbf{overlap-based} metrics, specifically \texttt{ROUGE} \cite{lin2004rouge} \texttt{METEOR} \cite{banerjee2005meteor} and  \texttt{BLEU} \cite{papineni2002bleu}. (2)~\textbf{similarity-based} metrics, that compute similarity between embeddings representations of generated and reference summaries. Specifically, we report \texttt{BERTScore} \cite{bertscore2020} and \texttt{MoverScore} \cite{zhao2019moverscore}. (3) a \textbf{QA-based} metric, specifically QAEval \cite{deutsch2021towards}. Although most QA-metrics are reference-free (discussed in Section \ref{sec:reference-free}), QAEval uses the reference summaries to indicate saliency. We report both exact match (EM) and F1 components of QAEval. 

\paragraph{Results} Table \ref{tab:reference-based-automatic} outlines the results. It shows that \brio~and \texttt{PEGASUS} models, fine-tuned to emulate the reference summaries, outperform \gpt~summaries according to all reference-based automatic metrics. The difference in their assigned scores is very high, e.g. >7 \texttt{ROUGE-L} points between \gpt~and \brio. For comparison, these reported scores for \gpt~are even lower than the trivial Lead-3 baseline reported in prior work \cite{fabbri2021summeval, grusky2018newsroom}. This clearly demonstrates that \textbf{current automatic reference-based metrics cannot be used to reliably measure  summary quality under the prompting paradigm}.

Amongst prompting-based models, we observe that \tzero~summaries report better metric scores than \gpt~for all datasets except Newsroom. Interestingly, out of the four datasets evaluated here, Newsroom is the only one not used to train the \tzero~model. This further shows that access to dataset-specific reference summaries during training improves performance according to these metrics, rendering them unsuitable for evaluating prompt-based models.

\subsection{Reference-free metrics}
\label{sec:reference-free}

Next, we investigate whether current reference-free evaluation metrics reflect the human preference rankings between summarization systems, as observed in Section~\ref{sec:human-study}. Here, we study 2 categories of metrics: (1) \textbf{quality metrics}, specifically SUPERT \cite{gao2020supert}, which evaluates generated summaries against automatically identified salient sentences in the input, and BLANC \cite{vasilyev2020fill}, which evaluates summaries on language understanding tasks. We refer readers to the original papers for detailed explanation of these. (2) \textbf{factuality metrics}, that are evaluate whether generated summaries contain incorrect information with respect to the source article. We report the performance of summarization systems using two QA-based metrics: QuestEval \cite{scialom2021questeval} and QAFactEval \cite{fabbri-etal-2022-qafacteval}. Additionally, we also benchmark entailment-based metrics: FactCC \cite{kryscinski2020evaluating}, DAE \cite{goyal2020evaluating, goyal2021annotating} and SummaC \cite{laban-etal-2022-summac}.\footnote{Exact model versions and configurations used for these are outlined in Appendix \ref{app:config}.} These entailment-based models are designed for classification into factual or non-factual; therefore, we use $P(\mathrm{factual \mid article, summary})$ to score generated summaries.

\paragraph{Results} Table \ref{tab:reference-free} outlines the scores for each summarization system according to the above reference-free metrics. Ideally, we want the relative rankings of different systems according to these metrics to correspond to human preferences, i.e. \gpt{} > \brio{} > \tzero{} for \texttt{CNN/DM}\footnote{Although the human study in Section~\ref{sec:human-study} is only run on CNN articles, the underlying fine-tuned model is same for both CNN and DM. Therefore, it we can reasonably expect it to display similar quality differences with respect to \gpt.} and \gpt{} > \tzero{} > \brio{} for \texttt{XSum}.\footnote{Note that while annotators were not explicitly asked to rate factuality, we instructed them to carefully check factuality and appropriately downvote non-factual summaries.}

Overall, we observe that none of the reference-free metrics we evaluate follow these trends for both \texttt{CNN/DM} and \texttt{XSum} datasets. In particular, we observe that \gpt{} summaries report low factuality scores (except \texttt{XSum}) even though we rarely found any factual errors in our qualitative analysis of its generated summaries. 

Interestingly, we noticed a roughly inverse relation to abstractiveness; summarization systems that generated more abstractive summaries (see Table \ref{tab:quantitative}) were generally scored lower by all automatic reference-based metrics. For instance, \gpt{} is scored lower than \brio{} by both quality metrics for all datasets except \texttt{XSum}; the latter is the only dataset for which \gpt{} summaries are less abstractive. Such shortcomings of reference-free evaluation metrics due to spurious correlations have also been studied in prior work \cite{durmus2022spurious}. These issues become more exaggerated when the summarization systems being compared exhibit very different properties.

\paragraph{Discussion} On the surface, the failure of reference-free metrics at evaluating \gpt{} summaries is more surprising that reference-based metrics as the later explicitly compares generated summaries with references that \gpt{} is not trained to imitate. Therefore, \gpt{} understandably scores lower than fine-tuned systems. 

However, we note two different issues with reference-free metrics: (1) Some of these, e.g. FactCC and DAE, use reference summaries as positive examples to train the metric. Therefore, although ``reference-free'' at test time, they are still trained to reward the summary properties seen in the standard summarization benchmarks. (2) Even completely reference-free metrics, e.g. QuestEval and QAFactEval, have only been evaluated on reference-based benchmarks and fine-tuned models. Therefore, the choice of different components, such as question answering or question generation models to use, etc. has been dictated by the error space of prior fine-tuned models \cite{tang2022understanding}. These decisions also now need to be re-visited to incorporate \gpt{} evaluation; we leave this for future work.

\section{Beyond Generic Summarization}
\label{sec:keyword}
Previously, we observed that \gpt{} models faithfully follow simple ``style'' instructions in the given prompts. This provides a promising direction to tackle other use cases in news summarization beyond the generic summarization task from Section~\ref{sec:human-study}.

\begin{figure*}[t]
    \centering
    \includegraphics[scale=0.30, trim=5mm 40mm 0mm 0mm, clip]{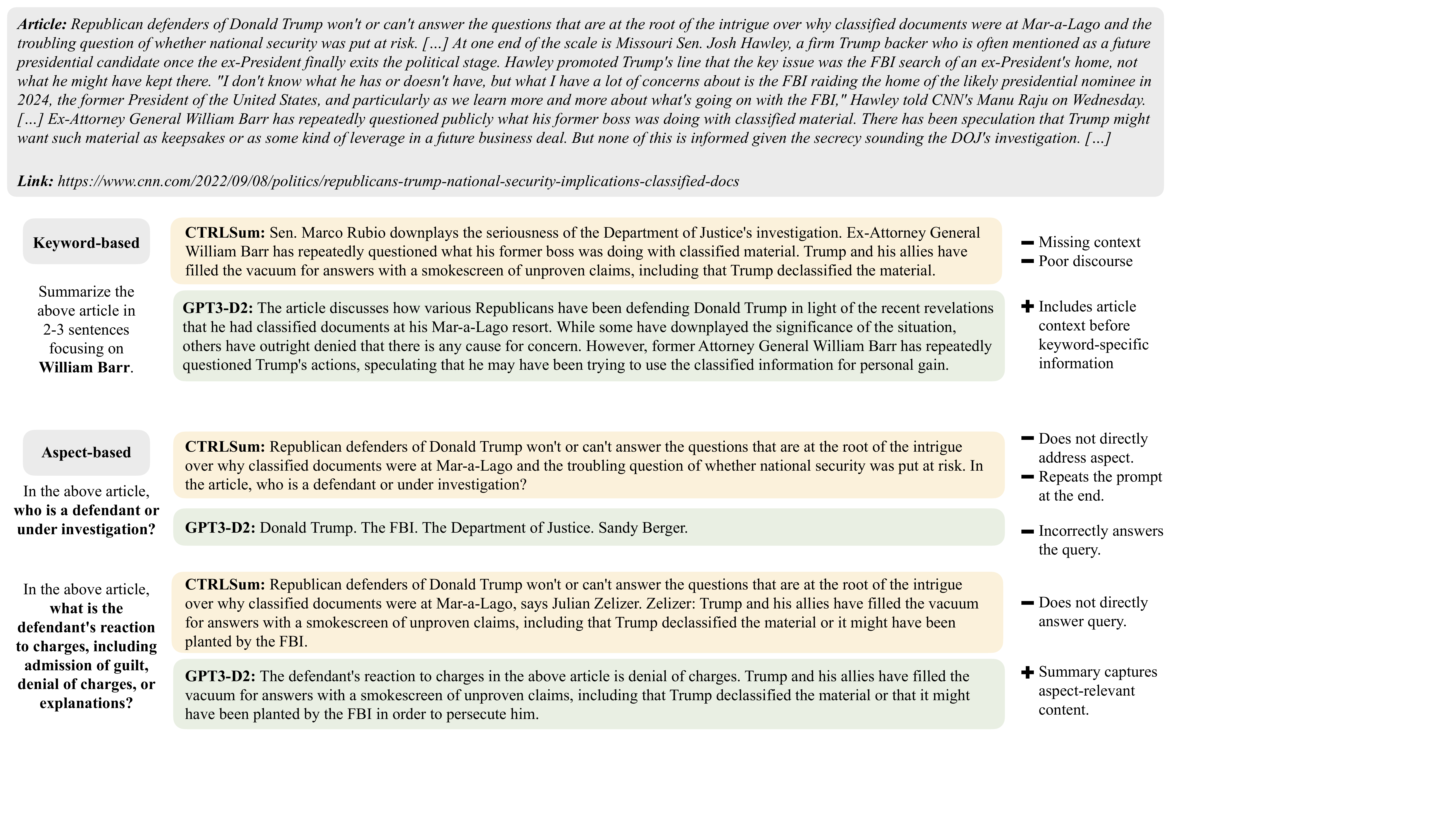}
    \caption{Comparison of keyword- and aspect-based summaries using \gpt~and \texttt{CTRLSum} models. The \gpt~prompt is shown on the left with the corresponding keyword or aspect bolded. For keyword-based summarization, the \gpt~summary presents appropriate context before the keyword-specific information. However, for aspect-based summarization, it does not always generate factually correct summaries, as shown in the first aspect example. We observe that \texttt{CTRLSum} performs poorly for both these settings.}
    \label{fig:keyword_focused}
\end{figure*}

Different users can have very different information needs from the same article, all of which cannot be satisfied with a single generic summary. Prior work has introduced several task formulations to address this gap, including keyword-focused \cite{he2020ctrlsum}, query-focused \cite{baumel2014query, he2020ctrlsum}, or aspect-focused summarization \cite{krishna2018generating, ahuja2022aspectnews}, amongst others. Here, we evaluate \gpt{} performance at two of these use cases.

In \textbf{keyword-based summarization}, the output summaries must succinctly summarize the input document focusing on a given keyword; these generally correspond to specific entities or events directly mentioned in the document. In contrast, the control units in \textbf{aspect-based summarization} are high-level topics that can be common across multiple similar types of documents. For e.g., for the input article in Figure~\ref{fig:length-keyword}, \textit{Donald Trump} or \textit{Russian interference in 2016 elections} are keyword controls whereas \textit{charges against the defendants} is a higher-level aspect that can serve as the query for any news article discussing a lawsuit or investigation.

\subsection{Qualitative Analysis}

\paragraph{Baseline Model for comparison} We use the recently proposed \texttt{CTRLSum} \cite{he2020ctrlsum}, a fine-tuned \texttt{BART} model, as our baseline. It can be flexibly adapted for both keyword- and aspect-based settings by including a prompt as additional input to the encoder. We use the prompt template recommended in the original paper.\footnote{Trained model publicly released at: \url{https://github.com/salesforce/ctrl-sum}.} 

\paragraph{Control Units} For the keyword-focused setting, we use named entities extracted from the input article as the control units. For aspect-focused summarization, we directly use the aspects introduced in the guided summarization task from TAC 2011.\footnote{\url{https://tac.nist.gov/2011/Summarization/Guided-
Summ.2011.guidelines.html}} It defined 5 broad categories of newswire articles, such as accidents and natural disasters, investigations and trial, etc., and multiple aspects for each category. For example, the \textit{``investigations and trials''} category includes aspects such as \textit{``who is the defendant or under trial?''}, \textit{``who is investigating, prosecuting, judging?''}, and so on. 

\paragraph{Qualitative Analysis} Figure~\ref{fig:keyword_focused} shows examples of keyword- and aspect-focused summaries using \gpt~and the baseline \texttt{CTRLSum} model. The keywords or aspects are highlighted in bold within the \gpt~prompt displayed on the left.   

In this example, representative of average \gpt{} quality, the keyword-focused \gpt~summary first gives a brief overview of the article setting before providing keyword-relevant information. In contrast, the  \texttt{CTRLSum} summary exhibits poor discourse structure and reads like a list of facts stapled together.  

The figure also shows aspect-focused summaries for two aspects associated with the ``investigations and trial'' category most appropriate for the chosen article. We see mixed results here for \gpt; it generates a factually incorrect summary for the first aspect, listing multiple people from the input article as defendants instead of only ``Donald Trump''. For the second aspect, it correctly maps the high-level concept ``defendant'' to ``Donald Trump'' in the input article and generates the correct answer to the input query: \textit{``The defendant's reaction to charges in the above article is denial of charges''}. 

On the other hand, \texttt{CTRLSum} fails to generate aspect-focused summaries for both cases. We believe that it struggles to align high-level concepts and explicit entities in the article due to a lack of such aspect-specific examples in its training data. Instead, it generates summaries focusing on lexically similar words, i.e. ``defenders'' for both cases.

Based off of \gpt's promising keyword-focused summarization capabilities observed above, we next conduct a human study to systematically compare it against the \texttt{CTRLSum} baseline. We leave further explorations of aspect-based summarization to future work, given the mixed to poor results for both models at this task.

\subsection{Human Study: Keyword-focused summarization}

\paragraph{Task Setup} Similar to Section~\ref{sec:human-study}, we design an A/B test to compare the two models. We use the same set of 100 CNN\footnote{We run this study using only CNN articles as the baseline \texttt{CTRLSum} model is trained on \texttt{CNN/DM}.} articles as Section~\ref{sec:human-study}. We randomly extract 2 distinct named entities from each article. In the study interface, the annotator is shown the article-keyword pair and \gpt{} and \texttt{CTRLSum} summaries corresponding to it. They are asked to select the summary that best summarizes the input article while focusing on the given keyword. Exact task instructions are included in Appendix \ref{app:instructions}.

Again, we run this study using the Prolific platform. We recruit 60 participants to annotate the 100 articles; each article is annotated by 3 annotators which includes annotations for 2 separate keywords. Each annotator evaluates 5 articles.

\begin{figure}[t]
    \centering
    \includegraphics[scale=0.24, trim=60mm 130mm 0mm 100mm, clip]{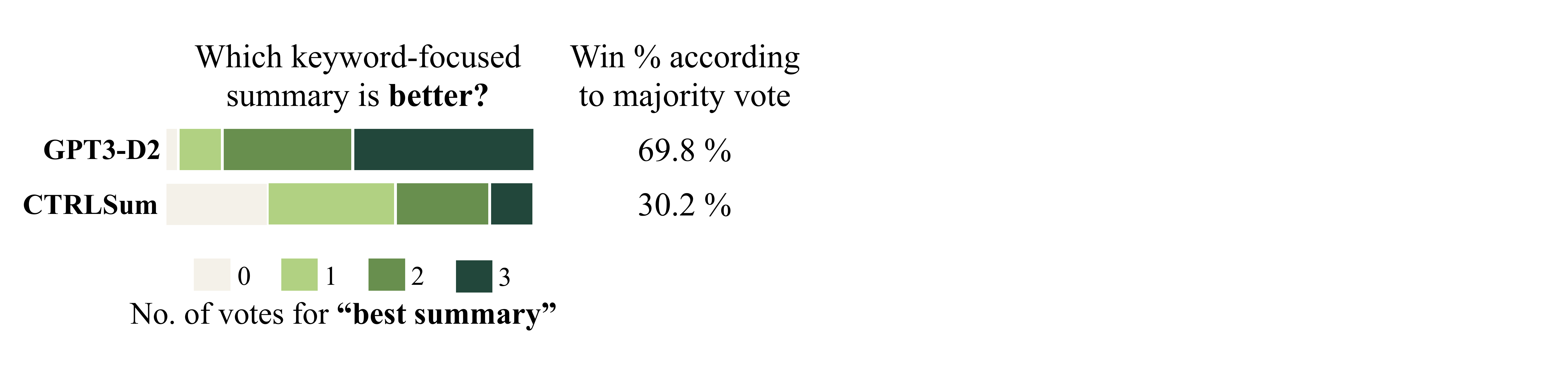}
    \caption{Distribution of annotator votes for the keyword-focused summarization task. Annotators prefer \gpt{} summaries over \texttt{CTRLSum} for approximately 70\% of all article-keyword pairs, showing unanimous preference more than half the time. }
    \label{fig:keyword}
\end{figure}

\paragraph{Results} Figure \ref{fig:keyword} shows the distribution of annotator votes between the \gpt{} and \texttt{CTRLSum} models. Annotators show a clear preference for \gpt{}. In fact, for nearly 70\% of all article-keyword pairs, \gpt{} is preferred over \texttt{CTRLSum} by a majority of the annotators. The main rationales given for this choice were better contextualization of keyword-related information and better coherence in \gpt{} summaries.

\paragraph{Impact} These results show that prompting \texttt{GPT-3} models present a promising alternative to fine-tuned models for such specialized summarization tasks that can be easily described using textual prompts. One of the major drawbacks of fine-tuned models is that they are constrained by what data is available and how it can be transformed to create new task-specific training data. \texttt{CTRLSum} relied on the SQuAD question answering dataset \cite{rajpurkar2016squad} because the required ``queries'' or ``questions'' were unavailable at scale for summaries in standard summarization datasets. In contrast, prompt-based models are not constrained by the availability of task-specific data and can flexibly adapt to new tasks. Future research should focus on further exploring these capabilities and possible improvements on currently ``unsolved'' tasks such as aspect-based or plan-based summarization.  

\section{Discussion and Related Work}
In recent years, research in text summarization  \cite{rush2015neural, nallapati2016abstractive, see2017get, lewis2020bart,zhang2020pegasus, liu2022brio} has typically relied on comparisons with gold test sets for evaluation, possibly augmented with reference-free metrics for dimensions like factuality. This paper shows that \textbf{all these metrics are completely ineffective at evaluating GPT-3 summaries}. Although issues with these metrics, particularly low correlation with human judgments, have also been studied earlier \cite{fabbri2021summeval, deutsch2021understanding}, they are considered reliable when comparing systems in different score ranges \cite{peyrard2019studying, deutsch-etal-2022-examining}. However, \texttt{GPT-3} challenges these established practices and evaluation protocols, and poses an urgent need for better evaluation.

This brings us to manual evaluation, generally considered to be the gold standard for generation evaluation. The majority of summarization research now reports results from a human study in addition to automatic metrics, but  there is a general lack of consensus on what dimensions to evaluate, task design, and other factors \cite{hardy2019highres}. This presents difficulties in conducting reliable and reproducible comparisons between systems \cite{karpinska2021perils}, another factor contributing to the popularity of automatic metrics. Although recent efforts like \texttt{GENIE} \cite{Khashabi2021GENIEAL} have taken steps to standardize manual evaluation protocols across systems, its annotation is not universally affordable and the quality is not strictly monitored. We hope that future work addresses these challenges and democratizes human evaluations.

The ultimate test of summarization systems is with actual users using the systems in practice. \citet{jones2007automatic} discusses the need to align task formulations with actual applications scenarios (``purpose factors''). However, the research in text summarization until now has been constrained to certain problems or domains by the heavy dependence on large-scale training data: for example, producing a bullet-point summary of a news article has emerged as standard due to availability of data from CNN, not because it is shown to be the best way to present information.

Now, the success of prompt-based models can allow realistic use-cases to drive research in a more top-down way. We already show that \gpt{} improves upon prior keyword-focused summarization systems that were trained on artificially adapted training data. In future research, we are interested in tackling other real world use cases, such as update summarization and plan- or aspect-based summarization. Additionally, adapting \gpt{} to documents longer than the allowed context, or structured inputs such as tables, presents research challenges beyond the current capabilities of \texttt{GPT-3} and would be interesting to study.\footnote{We very briefly discuss long document summarization with GPT-3 in Appendix \ref{app:long-doc}.}

\section{Conclusion}
In this work, we performed the first systematic study comparing prompt-based \texttt{GPT-3}  and fine-tuned models at the news summarization task. We analyzed the impact of prompting on the summarization field, including training paradigms and evaluation practices. Finally, to support further research in this direction, we release a large corpus of generated summaries for multiple prompt-based and fine-tuned models, as well as human preference judgments comparing these systems.

\section{Limitations}
In the text generation evaluation literature, there does not exist a standardized task design for comparing different system generations. In our work, we chose a human evaluation workflow that directly asks annotators to compare systems, while other prior work has opted for Likert-scale judgments and/or evaluation along multiple quality dimensions \cite{gehrmann2022repairing}. The latter strategy of evaluating different dimensions could surface more insights into which ``style'' properties of GPT-3 summaries provide them an edge over fine-tuned models; however, such analysis is outside the scope of this paper. Our experiments comparing overall quality  reveal that current summarization datasets are not well-aligned with user preferences. We leave more fine-grained analysis into these preference judgments for future work.

The experiments in this paper are run on English-language news summarization datasets as these serve as common benchmarks in the summarization literature. However, user rankings of system outputs might be different when evaluating other domains, e.g., summaries of scientific text. While we believe that automatic metrics would fail to evaluate GPT-3 summaries on these domains also (generated summaries would still look different from the reference summaries), users may prefer models that are specifically fine-tuned on domain-specific data for niche domains.

Finally, we do not know exact datasets or tasks used to train \gpt{}. It is possible that its RLHF training \cite{ouyang2022training} included summarization examples, and therefore, preference judgments from human annotators for its different outputs. However, our arguments in this paper do not rely on the specifics of the \gpt{} system, merely that such a system exists. If anything, the existence of potentially better data underscores that further work should collect new data for summarization model tuning, and our claims about metrics still hold regardless of the details of how the \gpt{} summaries were produced.

\bibliography{anthology,custom}
\bibliographystyle{acl_natbib}

\appendix

\section{Implementation Details}
\label{app:config}

\paragraph{Prompts Used} To generate \gpt{} summaries for all experiments in this paper, we use the standard prompt format outlined in Section~\ref{sec:background}. We set $N=3$ for CNN and DailyMail, $N=2$ for Newsroom, and $N=1$ for \texttt{XSum/BBC}. For the latter, the prompt is slightly modified to \textit{``Summarize the above article briefly in 1 sentence.''}

For T0, we use the following prompts: a) \texttt{CNN/DM}: \textit{``Summarize the article below in 3 to 4 sentences?''}, b) Newsroom: \textit{``Summarize the article below in 2 to 3 sentences?''}, and c) \texttt{XSum/BBC}: \textit{``Summarize the article below in 1 sentence?''}

\paragraph{Factuality Metrics} In Section~\ref{sec:reference-free}, we evaluated several recently proposed factuality metrics. We note that multiple versions have been released for some of these models in recent years. Here, we specify the versions used in our experiments to ensure reproducibility of results:
\begin{enumerate}[leftmargin=*]
    \item \textbf{QuestEval}: We use version 0.2.4 of the questeval python package and report numbers using the precision-only setting. 
    \item \textbf{DAE}: We use the updated version of the DAE model trained for document-level factuality. Latest code and model released at \url{https://github.com/tagoyal/factuality-datasets}. 
    \item \textbf{SummaC}: We use the SummaC-Conv model (model\_name = `vitc') and sentence-level granularity in our experiments.
\end{enumerate}

\paragraph{Keyword-based data} For our keyword-based human study, we extracted two named entities per article, as discussed in Section~\ref{sec:keyword}. In practice, we constrained the first keyword to be lead-biased, i.e. it was extracted from the first three sentences of the article, and the second keyword was extracted from the remaining article. As CNN-based summarization models are generally lead-biased, this allowed us to benchmark models under both settings.

\section{Are annotator judgments of quality correlated with length?} 
\label{app:length}

\begin{figure}[t]
    \centering
    \includegraphics[scale=.5, trim=0mm 0mm 0mm 10mm, clip]{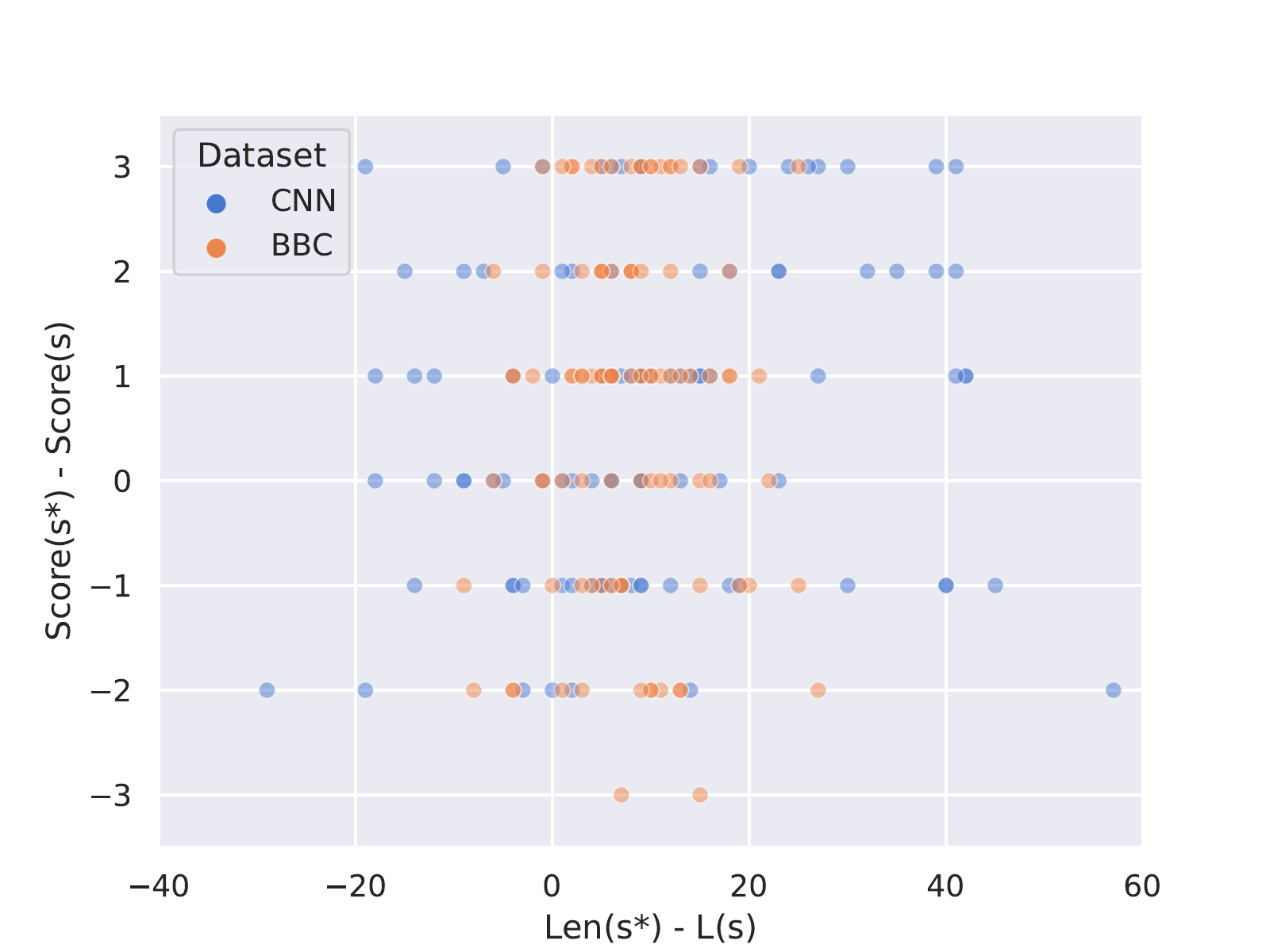}
    \caption{Correlation between summary length and annotator score (computed as the no. of ``best summary'' votes. For each example, plot the difference in length (x-axis) and annotator score (y-axis) between the \gpt{} summary and the next best system's summary.}
    \label{fig:length-corr}
\end{figure}

In Section \ref{sec:human-study}, results of the human study showed that annotators provide shorter length as one of the main reasons for selecting \tzero{} summaries as the worst for the CNN dataset. Here, we investigate if the choice between \gpt{} and \brio{} is similarly influenced by their length differences; \gpt{} summaries are on average 9 words longer. 

To study this, we plot the difference in summary length against the difference in annotator score (measured as the no. of votes for a summarization system) between the best summarization system (\gpt{}) and the next best system (\brio{} for CNN and \tzero{} for BBC). The resulting plot is shown in Figure~\ref{fig:length-corr}. In general we observe low correlation between these; Pearson's $\rho$ is 0.17 for CNN and .02 for the BBC dataset. These correlation values cannot solely explain the large differences in annotator judgments reported in the human study results of Section~\ref{sec:human-study}; additional quality factors must have influenced this choice. Anecdotally, we observe that the GPT summaries are slightly less information dense; our impression is that these contain a similar level of information content, but are easier to read and understand despite being a bit more verbose.

\section{Qualitative differences between GPT-3 versions}
\label{app:d1vsd2}

\begin{figure}[t]
    \centering
    \includegraphics[scale=0.25, trim=0mm 65mm 0mm 0mm, clip]{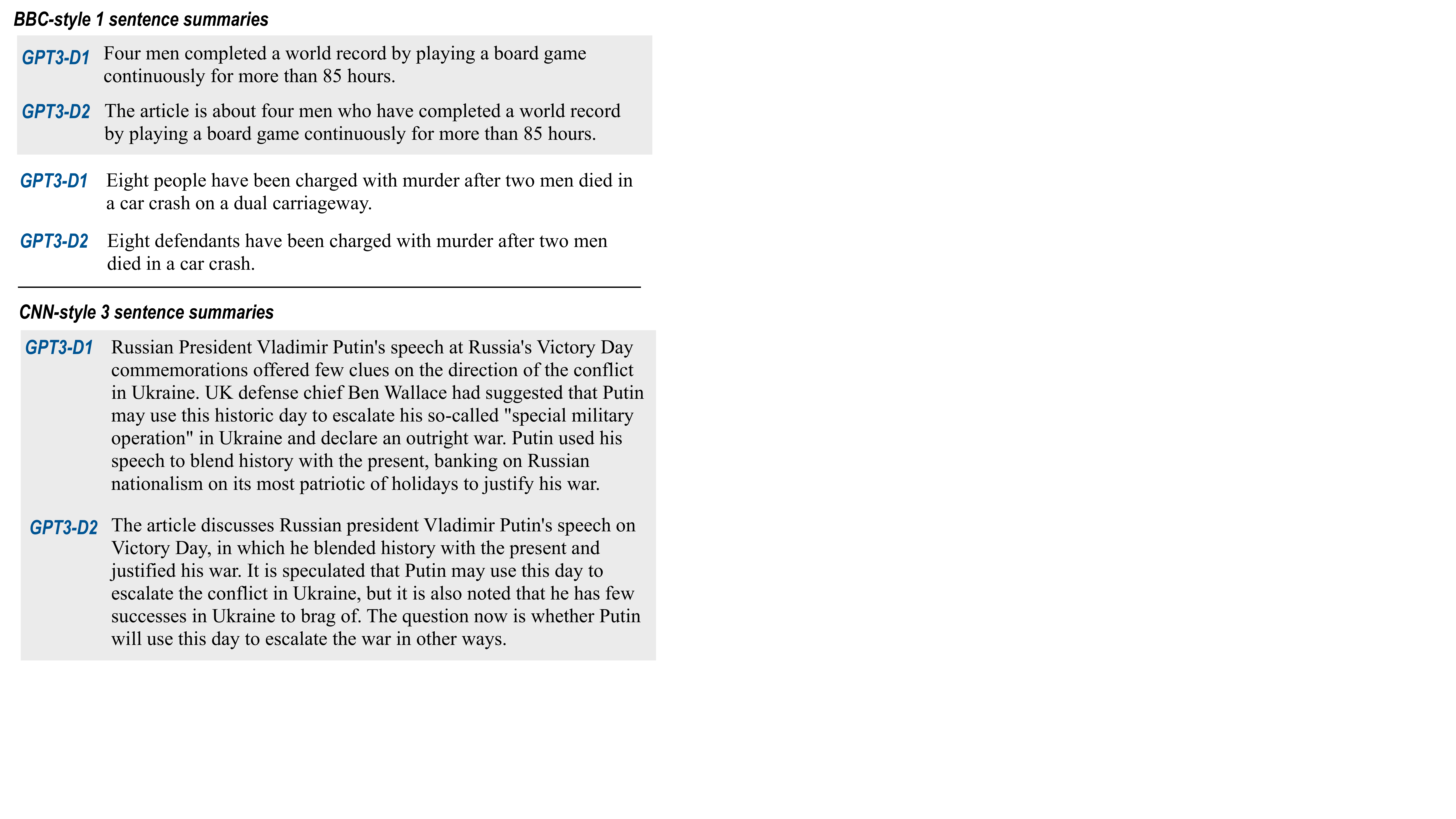}
    \caption{Examples of generated summaries using the text-davinci-001 (\texttt{GPT3-D1}) and text-davinci-002 (\gpt) versions. The figure shows both BBC and CNN-style summaries.}
    \label{fig:d1vsd2}
\end{figure}

Figure~\ref{fig:d1vsd2} shows examples comparing summaries from text-davinci-001 (\texttt{GPT3-D1}) to those from \gpt{}. For BBC-style single sentence summaries, we observed that the two models generated very similar summaries with high content and lexical overlap. More variance is observed for CNN-style summaries. In our anecdotal assessment, \texttt{GPT3-D1} generated more detailed summaries while those from \gpt{} are less information dense. 

\section{Human study and API costs}
\label{sec:cost}
  
At the time of running our experiments, \texttt{GPT-3} API's text-davinci-002 version was priced at \$0.06 per 1K tokens. New pricing information is available at: \url{https://openai.com/api/pricing/}.

In our experiments, we generated around 2600 \gpt{} summaries across all experiments in Section~\ref{sec:human-study} (human study), Section~\ref{sec:automatic-eval} (evaluation of metrics) and Section~\ref{sec:keyword} (keyword-based human study). We spent a total of approximately \$150 on API requests.

For the human study, we paid participants \$4 per task (each task involved annotation for 5 articles). On average, this translated to \$11/hr of work. The combined cost for the generic summarization (Section~\ref{sec:human-study})  and the keyword-based summarization (Section~\ref{sec:keyword}) studies was \$1020, including platform costs and bonus payments.

\section{Long document summarization using \gpt{}}
\label{app:long-doc}

Summarization of long documents has attracted significant interest in recent years \cite{cohan2018discourse, kryscinski2021booksum}. Here, we study how naive prompting of \texttt{GPT-3} performs at long-document summarization.

\begin{figure}[t]
    \centering
    \includegraphics[scale=0.27, trim=0mm 150mm 0mm 0mm, clip]{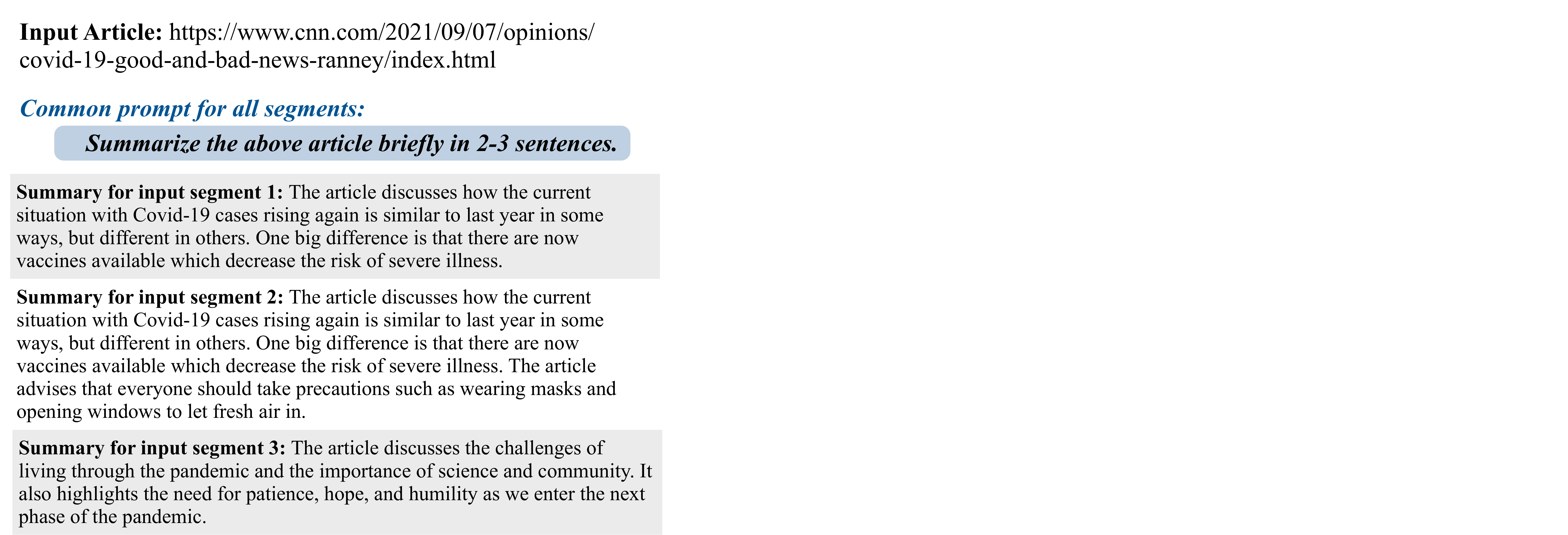}
    \caption{Illustrative example of \gpt{} summary of a long source article generated using the segment-then-summarize pipeline. }
    \label{fig:long-doc}
\end{figure}

First, we extract text from a long input article from the CNN website.\footnote{Article link: \url{https://www.cnn.com/2021/09/07/opinions/covid-19-good-and-bad-news-ranney/index.html}} Next, we follow the commonly used segment-then-summarize procedure from prior work \cite{zhao2020seal, zhang2022summn}. We divide the input article into 3 disjoint segments, summarize each segment separately and concatenate these outputs to form the final summary. 

Figure \ref{fig:long-doc} shows the prompt used and the generated summaries for each segment. While individual segment summaries are high quality, we can see that the concatenated summary is not coherent and includes repeated ``introductory'' sentences outlining similar content. Related to this, it also does not cover all important aspects of the input article as a majority of its `length budget' is spent on a high-level overview. We also observed that the generated summaries for long documents often focus on less unimportant parts of the document,  e.g. \textit{``...everyone should take the precaution of ... opening windows to let the fresh air in''} in the illustrated example. This is, in part, due to the segmentation of the input article: \gpt{} still exhibits some lead bias and treats the beginning of the input segment as more salient. Therefore, the exact segmentation of the article also dictates the quality of the final summary, and cannot be readily fixed by altering the prompt.

These observations show that while \gpt{} produces superior segment-level summaries, it is more difficult to adapt it to ``non-natural'' text inputs without fine-tuning. Therefore, techniques that have shown promising results for fine-tuned models, e.g. segment-then-summarize or extract-then-abstract \cite{zhang2021exploratory} approaches, are not as effective when directly applied with prompting-based models. 

\section{Task Instructions}
\label{app:instructions}

\begin{figure}[h]
    \centering
    \includegraphics[scale=.24, trim=5mm 22mm 0mm 0mm, clip]{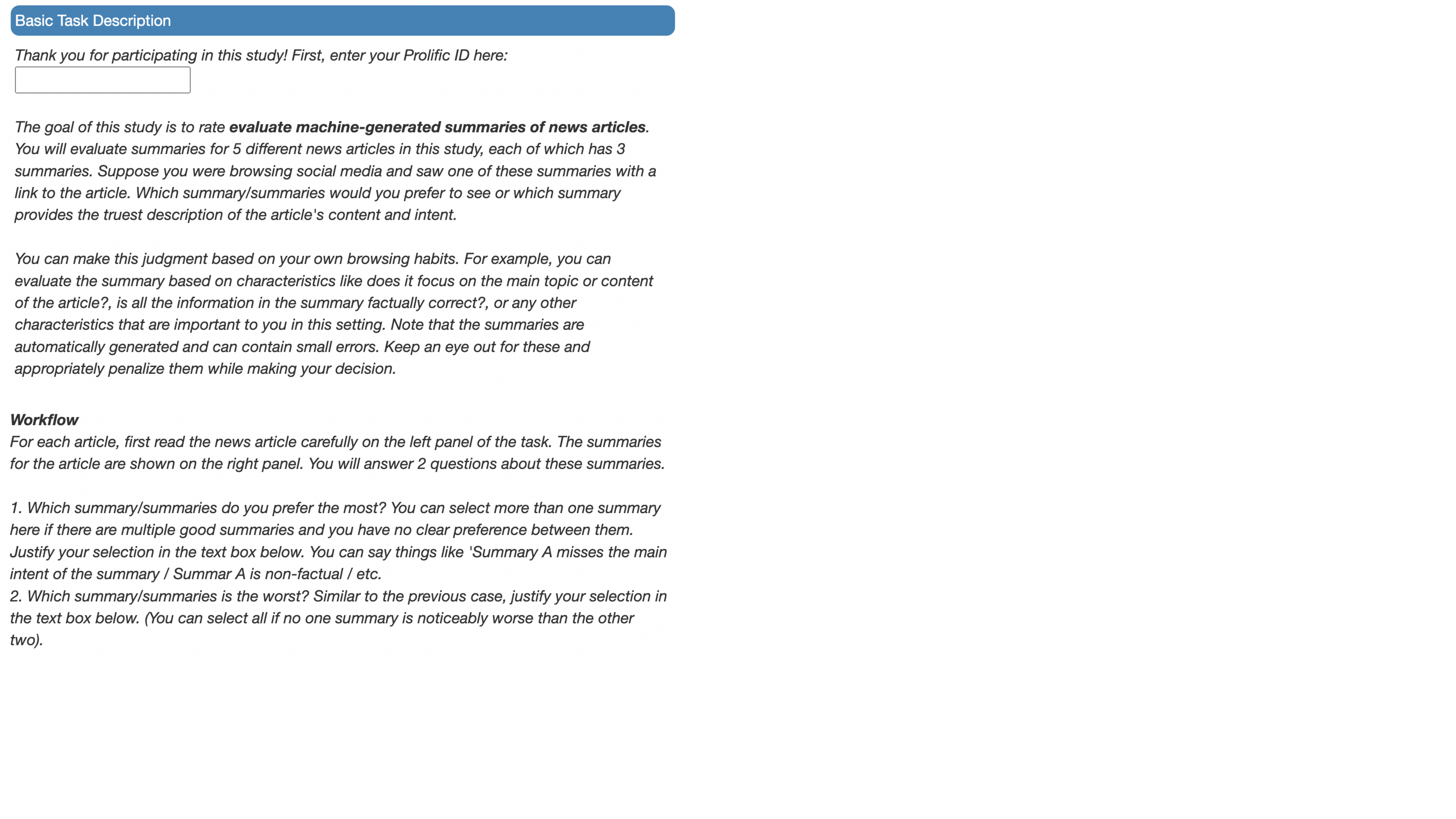}
    \caption{Screenshot of the task instructions for the generic summarization setting.}
    \label{fig:basic_instructions}
\end{figure}

\begin{figure}[t]
    \centering
    \includegraphics[scale=.27, trim=5mm 63mm 0mm 0mm, clip]{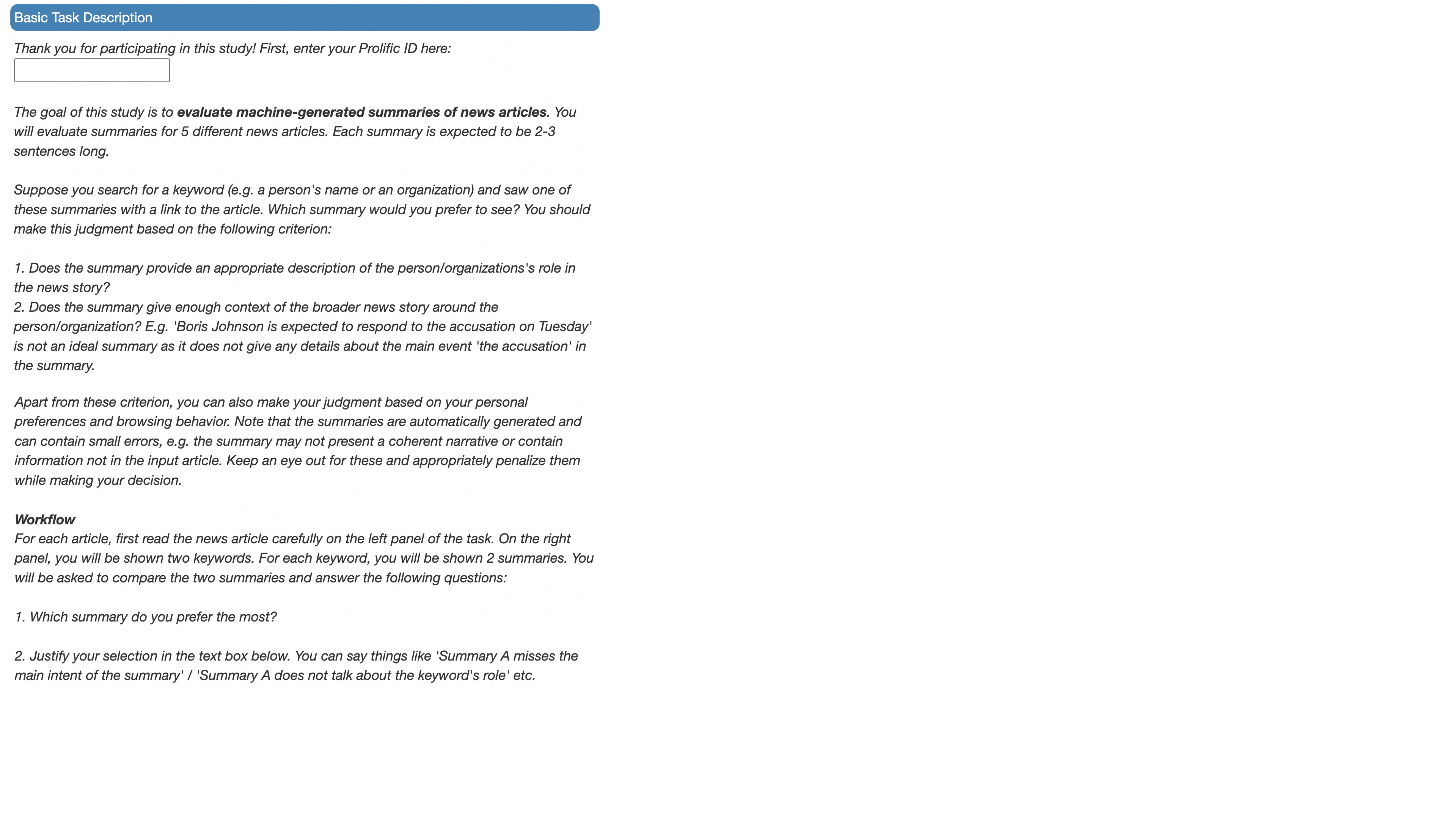}
    \caption{Screenshot of the task instructions for the keyword-based setting.}
    \label{fig:keyword_instructions}
\end{figure}

Task instructions provided to crowd annotators for the generic summarization task setting are shown in Figure~\ref{fig:basic_instructions} and those for the keyword-based setting are shown in Figure~\ref{fig:keyword_instructions}.

\section{Examples of generated summaries}
\label{app:examples}
\begin{figure*}
    \centering
    \includegraphics[scale=0.38,trim=0mm 0mm 0mm 0mm, clip]{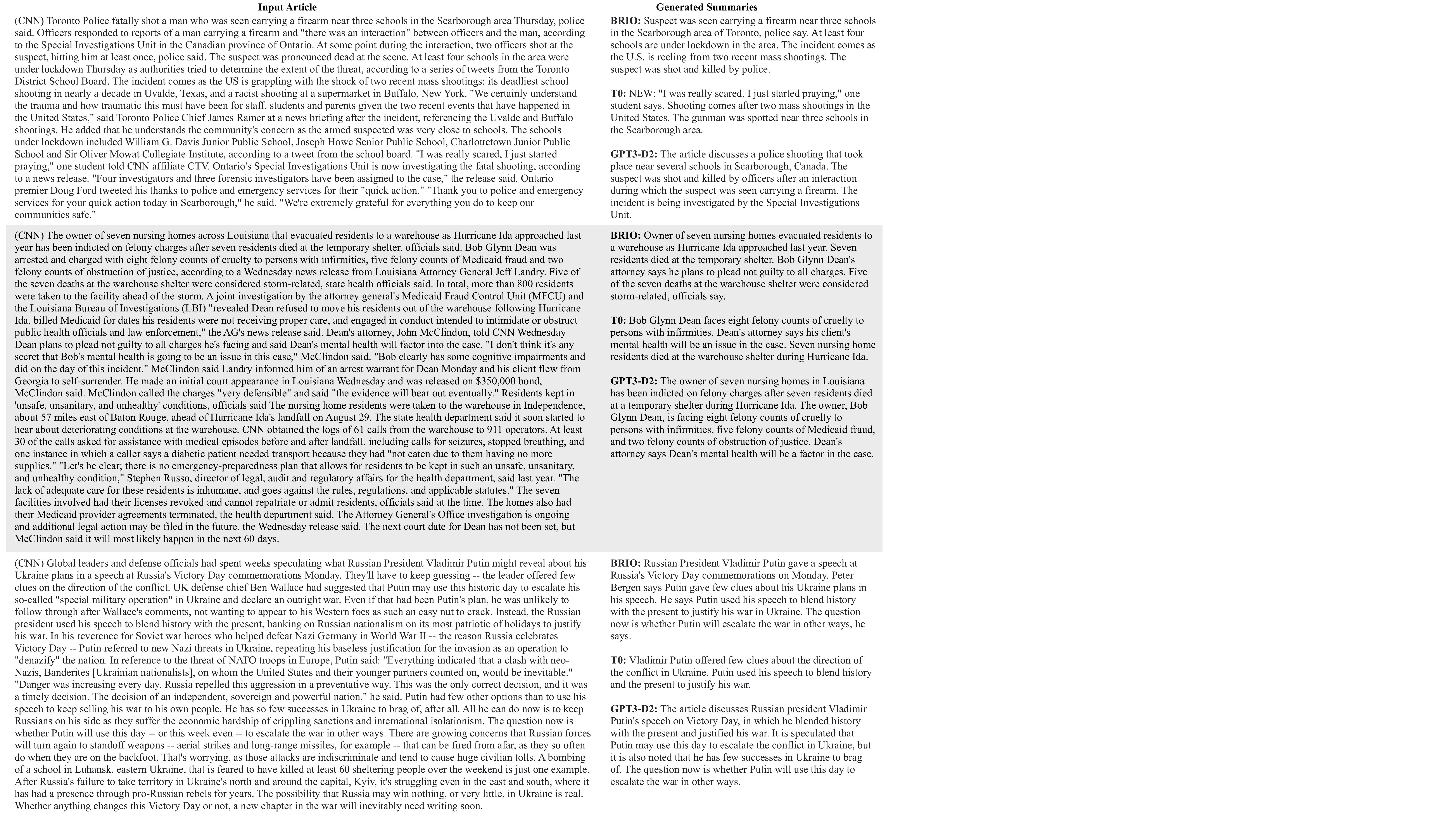}
    \caption{Examples of generated summaries for the CNN-2022 dataset using 3 different summarization systems.}
    \label{fig:appendix_cnn}
\end{figure*}

\begin{figure*}
    \centering
    \includegraphics[scale=0.38,trim=0mm 190mm 0mm 0mm, clip]{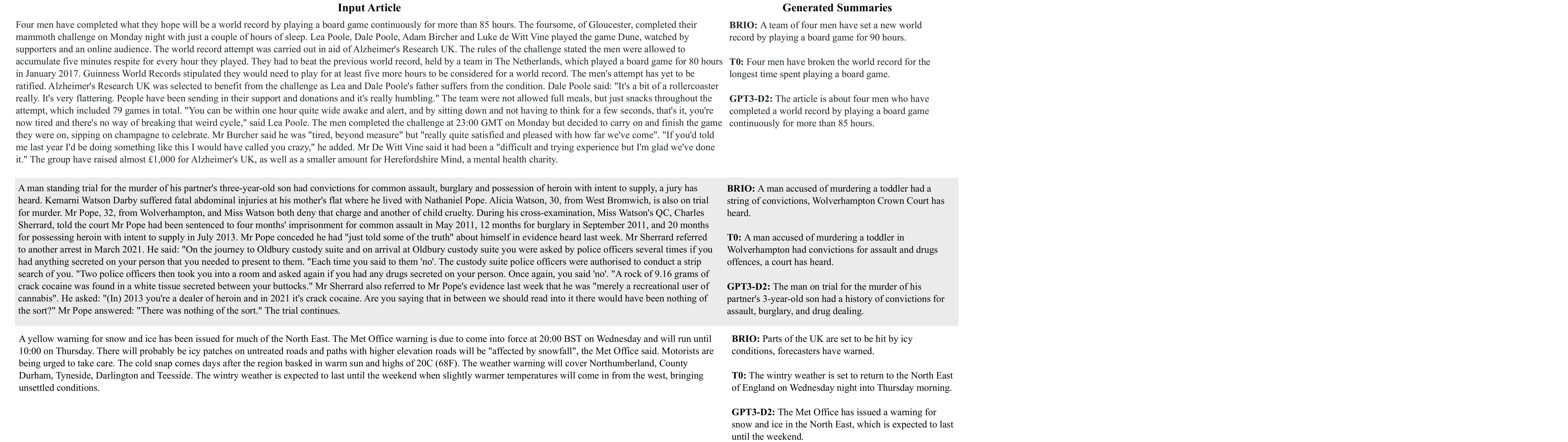}
    \caption{Examples of generated summaries for the BBC-2022 dataset using 3 different summarization systems.}
    \label{fig:appendix_bbc}
\end{figure*}

 We show examples of generated summaries for articles for generic summarization for CNN-2022 and BBC-2022 in Figures~\ref{fig:appendix_cnn} and \ref{fig:appendix_bbc}. It includes summaries from the 3 different summarization models evaluated in the human study in Section~\ref{sec:human-study}.
 
\begin{figure*}
    \centering
    \includegraphics[scale=0.38,trim=0mm 90mm 0mm 0mm, clip]{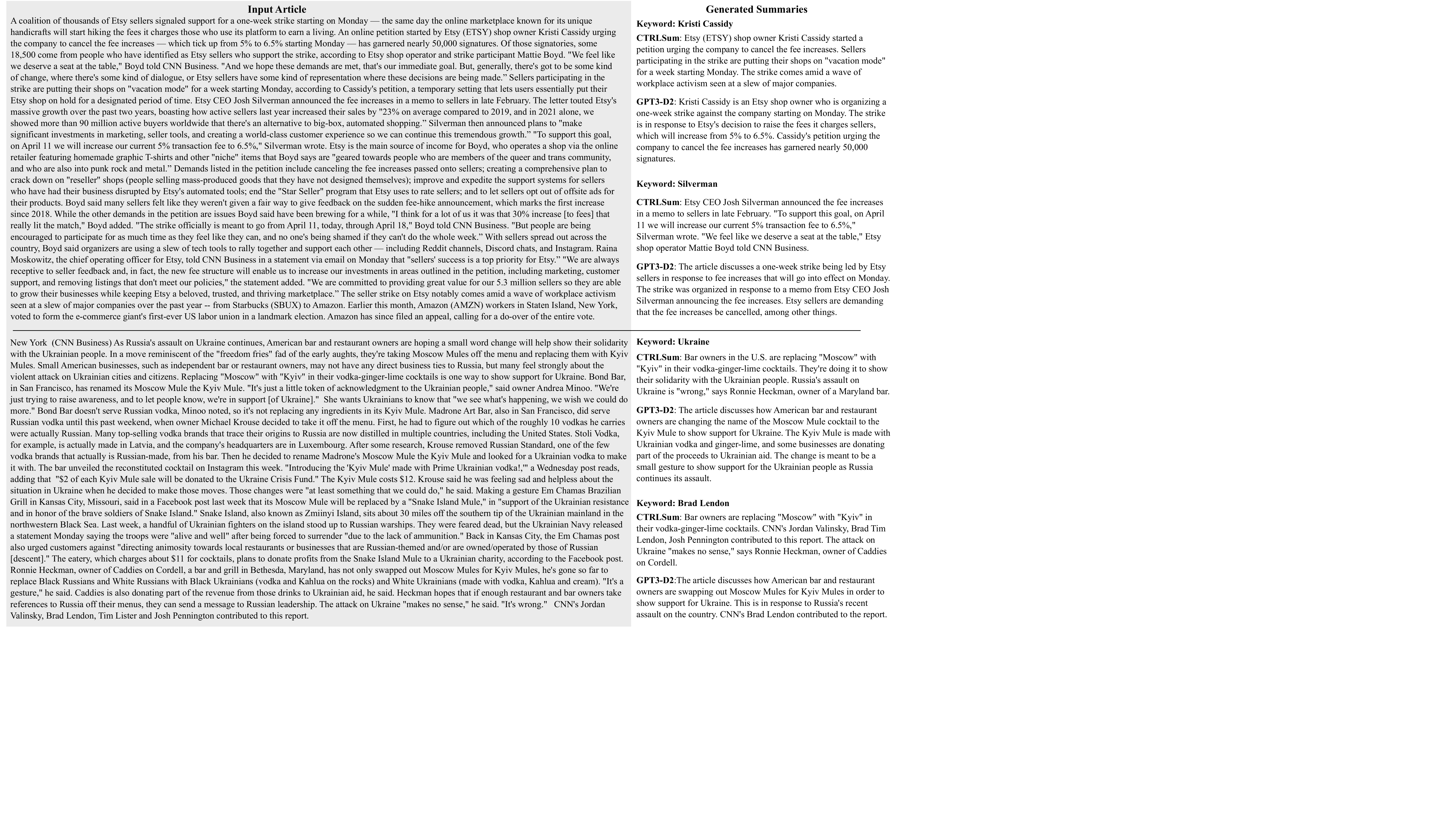}
    \caption{Examples of keyword-focused summaries for CNN articles from 2022.}
    \label{fig:appendix_keyword}
\end{figure*}

 Examples of keyword-focused summaries are shown in Figure~\ref{fig:appendix_keyword} for CNN. It includes summaries generated by \gpt{} and \texttt{CTRLSum} models.

\end{document}